\documentclass[11pt]{article}

\usepackage[margin=1in]{geometry}

\usepackage{amsmath}
\usepackage{amssymb}

\usepackage{graphicx}
\usepackage{booktabs}
\usepackage{multirow}
\usepackage{subfig}
\usepackage{float}

\usepackage{algorithm}
\usepackage{algorithmic}

\usepackage[hidelinks]{hyperref}

\title{Structural Inference under Hidden Agents}

\author{
Zhongben Gong \and
Xiaoqun Wu\thanks{Corresponding author: \texttt{xqwu@szu.edu.cn}} \and
Mingyang Zhou \and
Hui Huang
\\[0.5em]
\small College of Computer Science and Software Engineering, Shenzhen University,\\
\small Shenzhen 518060, China
}

\date{September 2026}

\begin{document}

\maketitle

\begin{abstract}
Recovering latent interaction structures from multi-agent dynamics is important
for understanding and predicting interacting systems. Trajectory-based structural
inference has achieved promising performance, but conventional formulations assume
that the trajectories of all modeled agents are available. In practice, agents may
become unobserved at deployment because of limited sensing, occlusion, or
communication failure. Existing studies have considered unseen-node estimation,
structural inference under partial observations, and missing-value imputation, yet
the joint recovery of hidden-agent trajectories and their interactions remains
underexplored. We formulate this problem as structural inference under hidden agents.
Its key difficulty is a circular dependency: recovering interactions involving a
hidden agent requires an estimate of its trajectory, while trajectory reconstruction
can itself benefit from structural information. To address this challenge, we propose
Structural Inference under Hidden Agents (SIHA), which combines structure-agnostic
initialization with structure-guided iterative refinement. SIHA reconstructs hidden
trajectories from visible observations, infers interactions using Neural Relational
Inference, and feeds the estimated structure back into hidden-state reconstruction
through multi-strength structural attention and iterative state--structure updates.
Experiments on three benchmark dynamical systems demonstrate consistent improvements
in visible-to-visible structural inference, while also showing benefits in hidden-state
reconstruction and future prediction. Motion-capture experiments with simulated
whole-limb occlusion further demonstrate its effectiveness in realistic hidden-agent
settings.
\end{abstract}


\section{Introduction}

Interaction and network structure shape observed dynamics across many scientific domains. In physical systems, collective dynamics contain information about the underlying interaction network~\cite{nitzan2017revealing}. In biology, inferring gene regulatory networks from single-cell measurements helps characterize regulatory organization~\cite{pratapa2020benchmarking}. Social-network experiments demonstrate that network structure affects behavioral diffusion~\cite{centola2010spread}, while intersectoral production networks shape how shocks propagate into aggregate economic fluctuations~\cite{acemoglu2012network}. These examples motivate methods for understanding or recovering latent interactions from observations of collective behavior.

\begin{figure}[t]
    \centering
    \includegraphics[width=0.5\linewidth]{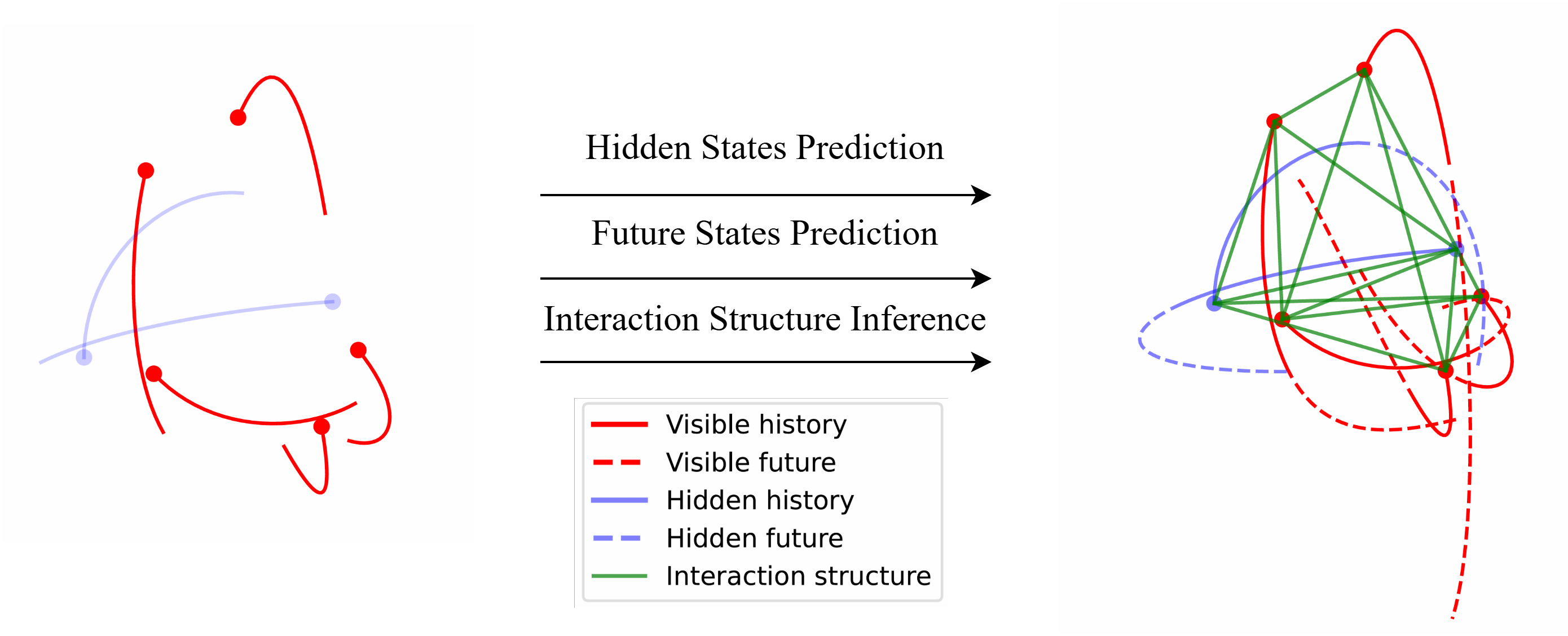}
    \caption{
    Illustration of the problem setting considered in this paper. 
    Given the observed trajectories of visible agents, 
    the goal is to predict the hidden agents’ trajectories, all future trajectories, and recover the latent interaction structure.
    }
    \label{fig:intro}
\end{figure}

Trajectory-based neural structural inference pursues this objective by learning relations among modeled entities from their state sequences. Neural Relational Inference (NRI)~\cite{kipf2018neural} provides a central formulation in which a latent interaction graph supports predictive dynamics; later extensions consider time-varying relations or iterative graph refinement~\cite{graber2020dynamic,wang2022iterative}. These conventional formulations typically operate on states or trajectories of the entities supplied as input. In practical systems, however, this assumption can be violated at the entity level. For example, in biological systems, the dynamics of some interacting species may remain unobserved because field surveys or sensors capture only a subset of the ecosystem; in engineered multi-agent systems, vehicles or robots may become unobservable because of occlusion, sensing limitations, or communication loss; and in motion-capture scenarios, some body parts may be entirely missing from a sequence because of persistent occlusion or tracking failure. In these cases, an interacting entity may be entirely absent from the observations over the considered time interval, rather than merely having sporadic missing values. Its state is therefore absent as a direct input, while interactions involving that entity must still be inferred through the dynamics of the observed agents. This information gap makes complete structural inference under-constrained, as illustrated in Figure~\ref{fig:intro}.

Neighboring work studies unseen-node inference, node-level partial observation, and relational inference with missing values or temporal segments~\cite{alet2019neural,wang2024conjoined,zheng2026diffusion}; these regimes differ in their observation units, inference targets, and supervision, as detailed in Section~2.2. We consider a protocol in which complete multi-agent trajectories are available during training: selected agents are masked from the predictor input, and their ground-truth trajectories supervise hidden-state reconstruction, while ground-truth edge labels are not used for model training. At deployment, only visible-agent trajectories are provided, and the latent interaction structure is inferred separately for each sample rather than assumed to be shared across different observations. SIHA therefore reconstructs the hidden-agent trajectories and estimates the corresponding sample-specific interactions involving hidden agents. Compared with conventional structural inference under fully observed trajectories or fixed missing-value settings, this setting remains comparatively underexplored. This protocol matches the practical scenarios described above. In controlled training environments, complete trajectories can be collected with reliable communication or multi-view sensing, whereas deployment may involve occlusion, signal loss, or limited sensing coverage that leaves some entities entirely unobserved. Moreover, interaction patterns can vary across missions, scenes, or motions, while explicit edge annotations may be unavailable, making sample-specific structural inference from the remaining visible trajectories a realistic requirement.

Within this protocol, hidden-state reconstruction and complete structural inference depend on one another. Structural information can guide the reconstruction of unobserved agents, whereas complete structural inference in the current SIHA data flow relies on reconstructed hidden trajectories. This coupled inference problem motivates a structure-agnostic state initialization followed by structure-guided, iterative refinement of states and interactions. SIHA follows this computational flow. A structure-agnostic hidden-state predictor (HSP-sa) first initializes the hidden trajectories from visible observations. A standard NRI model, used in the current implementation as the structure-inference and future-prediction backbone, then estimates interactions from the visible and reconstructed trajectories. A structure-guided hidden-state predictor (HSP-sg) refines the hidden trajectories using the inferred structure. Its multi-strength structural guidance retains learnable attention paths alongside attention biased by the predicted structure, after which HSP-sg and NRI alternate to refine the reconstructed states and structural estimate.

We evaluate SIHA on the Springs, Charged Particles, and Kuramoto systems with one to five hidden agents. In the principal external comparison, SIHA and visible-only NRI are compared numerically only on visible-agent forecasting and visible-to-visible structural accuracy, which are defined for both methods. SIHA obtains higher visible-to-visible structural accuracy throughout the reported grid, while visible-agent forecasting results vary by system. Hidden histories, hidden future trajectories, and interactions involving hidden agents represent additional SIHA outputs that visible-only NRI does not define under this protocol. The internal comparison contrasts the HSP-sa baseline with the complete SIHA pipeline: both use matched hidden-state supervision and the same pretrained NRI module, while SIHA further introduces HSP-sg for structure-guided reconstruction and iterative state--structure refinement. Across the reported settings, SIHA matches or improves upon HSP-sa on the displayed metrics. The evaluation further examines hidden-agent-count trends, recorded CMU Motion Capture trajectories with simulated whole-limb occlusion~\cite{CMUMocap2003}, and mechanism and supervision analyses.

The contributions of this work are threefold:
(1) we introduce structural inference under hidden agents, where complete trajectories can be available for training but only visible-agent trajectories are observed at deployment, and hidden trajectories and their interactions must be jointly recovered without edge-label supervision;
(2) we propose SIHA to resolve the circular dependency between hidden-state reconstruction and structural inference through structure-agnostic bootstrapping, structure-guided reconstruction, and iterative state--structure refinement; and
(3) we demonstrate the effectiveness of SIHA across three dynamical systems, varying numbers of hidden agents, and motion-capture sequences with whole-limb occlusion, showing consistent improvements in visible-to-visible structural inference and clear benefits from structure-guided refinement.

\section{Related Work}
\subsection{Relational and Structural Inference}
Recovering interactions from collective dynamics is a classical inverse problem~\cite{nitzan2017revealing}. Interaction Networks make objects and pairwise relations explicit when the graph is supplied~\cite{battaglia2016interaction}, whereas Neural Relational Inference (NRI) learns discrete latent edge types from trajectories through a predictive decoder without ground-truth edge labels~\cite{kipf2018neural}.

Subsequent work extends this paradigm through joint structure--dynamics learning and improved message passing~\cite{zhang2019general,chen2021neural}, time-varying or evolving relations~\cite{graber2020dynamic,li2020evolvegraph}, and iterative graph refinement~\cite{wang2022iterative}. Other formulations infer relations through masked reconstruction, heterogeneous interaction modeling, or learned discrete graph structures~\cite{grossmann2023masked,han2024collective,franceschi2019learning}. Parallel time-series studies also estimate causal or predictive dependencies among observed variables or variable groups~\cite{cai2024granger,wang2024credible}. Collectively, these methods broaden relational and structural inference while generally assuming that modeled entities are represented by observed states, trajectories, or node attributes. SIHA considers deployment in which entire agent trajectories are absent and must be reconstructed together with their interactions.

\subsection{Structural Inference under Incomplete Observations}
Incomplete observation arises at multiple levels, including unseen nodes, partially observed topology, scarce state samples, and masked values or temporal segments. Several studies address closely related incomplete-observation settings. Alet et al.~\cite{alet2019neural} estimate an unseen node at test time by optimizing its initial state under learned dynamics; their Section~5.2 demonstration uses a predictive model trained with ground-truth edges. SICSM~\cite{wang2024conjoined} studies structural inference under node-level partial observation and represents effects of hidden intermediaries through indirect, multi-hop dependencies. Its primary target is structure under partial observation, without jointly treating explicit unobserved-node trajectory reconstruction and complete hidden-incident graph recovery as the inference objective. DiffRI~\cite{zheng2026diffusion} combines self-supervised diffusion imputation with relational inference when values or temporal segments are masked within a fixed set of known components. It imputes component states, while an entirely absent component trajectory lies outside its input formulation.

Neighboring regimes include recovering network-generating rules from partially observed topology~\cite{yang2021hidden} and learning continuous network dynamics from sparse, irregular, partial, or noisy state observations~\cite{cui2024continuous}. These regimes differ in the missing unit and inference target. Within this taxonomy, SIHA jointly reconstructs whole hidden-agent trajectories and estimates interactions involving hidden agents. Complete trajectories provide hidden-state supervision during training without edge-label supervision, whereas deployment uses visible-agent trajectories only.

\subsection{State Reconstruction and Graph-Guided Imputation}
Time-series imputation reconstructs missing observations over a fixed set of known variables using recurrent, probabilistic, attention-based, or diffusion models, including BRITS, GP-VAE, SAITS, and CSDI~\cite{cao2018brits,fortuin2020gpvae,du2023saits,tashiro2021csdi}. GRIN further uses graph message passing to incorporate relational information into multivariate imputation~\cite{cini2022filling}. These methods show how temporal and relational context can support reconstruction of missing values or segments in known variables. Their primary objective remains value-level state reconstruction over fixed channels. SIHA addresses an agent absent as an entity from the deployment input and couples the two directions: estimated structure guides hidden-state reconstruction, while reconstructed hidden trajectories support inference of interactions involving hidden agents.

\section{Problem Formulation}
\label{sec:problem_formulation}

\subsection{Interacting Systems with Hidden Agents}
\label{sec:hidden_setting}

We consider an interacting dynamical system with $N$ agents, of which the first $N_{\text{vis}}$ agents are visible and the remaining $N_{\text{hid}}=N-N_{\text{vis}}$ agents are hidden. The state of agent $i$ at time $t$ is denoted by $\mathbf{x}_i^t\in\mathbb{R}^d$. Its trajectory over an observed history of $T$ time steps is
\begin{equation}
    \mathbf{x}_i=[\mathbf{x}_i^1,\ldots,\mathbf{x}_i^T]\in\mathbb{R}^{T\times d}.
\end{equation}
The visible- and hidden-agent trajectories are respectively collected as
\begin{equation}
\begin{aligned}
    \mathbf{x}_{\text{vis}}
    &= [\mathbf{x}_1,\ldots,\mathbf{x}_{N_{\text{vis}}}]
    \in\mathbb{R}^{N_{\text{vis}}\times T\times d},\\
    \mathbf{x}_{\text{hid}}
    &= [\mathbf{x}_{N_{\text{vis}}+1},\ldots,\mathbf{x}_N]
    \in\mathbb{R}^{N_{\text{hid}}\times T\times d}.
\end{aligned}
\end{equation}

The interaction structure among all agents is represented by a directed graph $\mathcal{G}=(\mathcal{V},\mathcal{E})$, where $\mathcal{V}=\{v_1,\ldots,v_N\}$ and $\mathcal{E}\subseteq\mathcal{V}\times\mathcal{V}$. For $K$ possible interaction types, the complete graph is encoded by an adjacency tensor $\mathbf{z}\in\mathbb{R}^{N\times N\times K}$, where $z_{ijk}=1$ indicates an interaction of type $k$ from agent $i$ to agent $j$. Because entire hidden-agent trajectories are absent from the observed input, both their states and their interactions with the rest of the system must be estimated.

\subsection{Training and Deployment Setting}

During the training phase, complete multi-agent trajectories are available. To train the hidden-state predictors, a designated subset of agents is masked from the predictor input to form $\mathbf{x}_{\text{vis}}$, while the corresponding ground-truth trajectories $\mathbf{x}_{\text{hid}}$ are used as reconstruction targets. The structure-inference module is pretrained on complete trajectories using the standard NRI objective, without using the ground-truth interaction tensor $\mathbf{z}$ as a training label. When ground-truth structures are available in the experimental datasets, they are used only for offline evaluation.

At deployment, only $\mathbf{x}_{\text{vis}}$ is observed. SIHA first reconstructs the hidden trajectories and then combines them with the visible trajectories for future-state prediction and complete-structure inference; ground-truth hidden states and interaction labels are not used during this process.

\subsection{Inference Targets}

Given the visible-agent histories $\mathbf{x}_{\text{vis}}$ at deployment, the task is to estimate three inference targets:

(1) the historical trajectories of the hidden agents,
$\hat{\mathbf{x}}_{\text{hid}}\in\mathbb{R}^{N_{\text{hid}}\times T\times d}$;

(2) the future trajectories of all agents over a horizon of $T'$ steps,
$\hat{\mathbf{x}}_{\text{all}}^{\text{future}}\in\mathbb{R}^{N\times T'\times d}$; and

(3) the complete interaction structure,
$\hat{\mathbf{z}}\in\mathbb{R}^{N\times N\times K}$, including interactions involving visible and hidden agents.

The overall inference problem can thus be written as
\begin{equation}
    f:\mathbf{x}_{\text{vis}}
    \mapsto
    \left(
    \hat{\mathbf{x}}_{\text{hid}},
    \hat{\mathbf{x}}_{\text{all}}^{\text{future}},
    \hat{\mathbf{z}}
    \right).
\end{equation}

\section{Method}
\label{sec:method}
\subsection{Framework Overview}
\label{sec:framework_overview}

The key challenge in structural inference with hidden agents is the circular dependency between hidden-state reconstruction and structural inference. Inferring interactions involving a hidden agent requires an estimate of its trajectory, while reconstructing that trajectory can benefit from knowing how the agent interacts with the observed system. At deployment, neither quantity is directly available, so the inference procedure must first establish an initial estimate before structural information can be exploited.

\begin{figure}[H]
    \centering
    \includegraphics[width=0.95\linewidth]{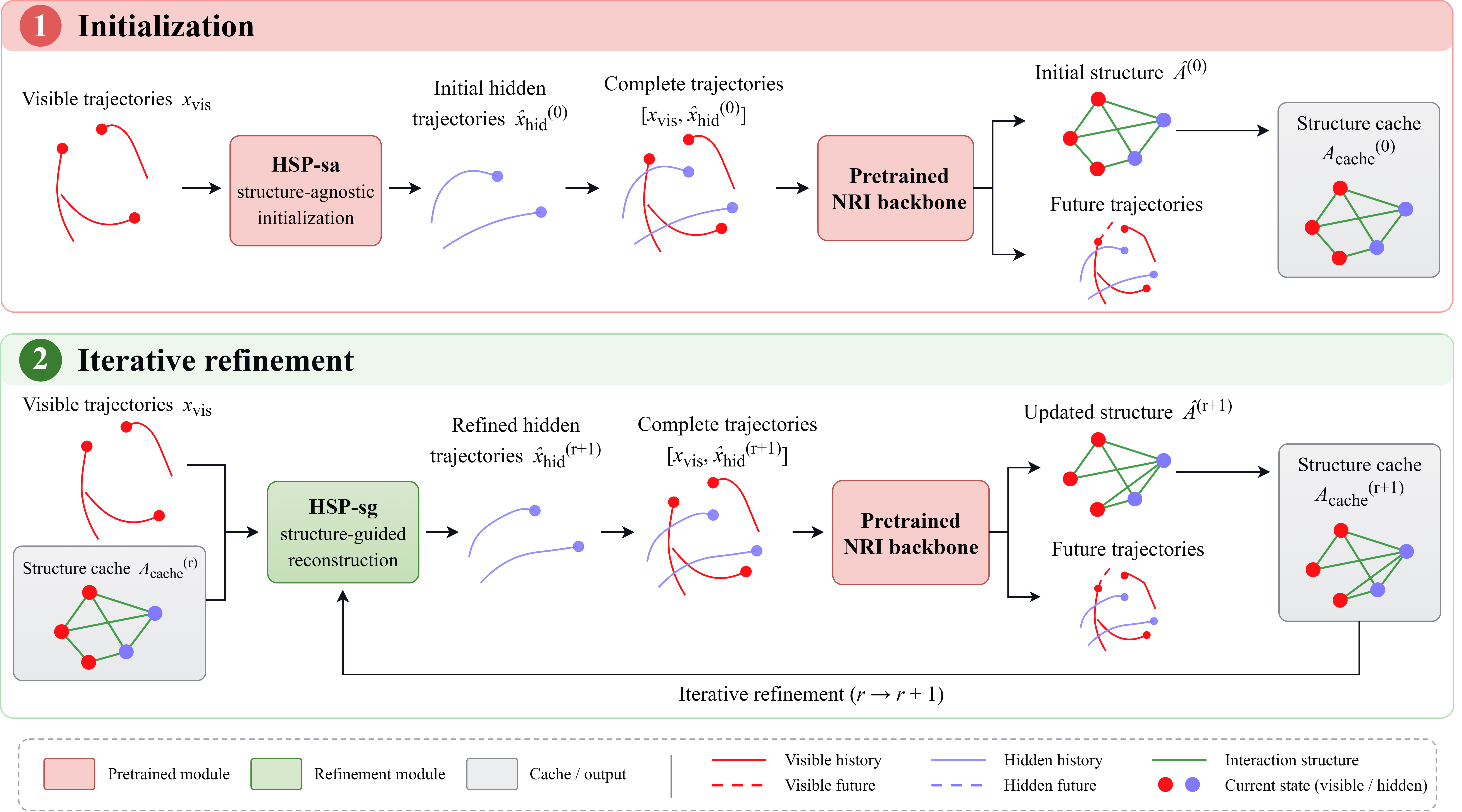}
    \caption{
       Overview of the SIHA state--structure refinement framework. HSP-sa initializes the hidden trajectories, after which the pretrained NRI backbone estimates the interaction structure and future trajectories from the completed trajectories. HSP-sg then uses the cached structure to refine the hidden trajectories, and NRI recomputes the structure for subsequent refinement. Purple boxes denote pretrained modules, while the green box denotes the structure-guided refinement module.
    }
    
    \label{fig:framework}
\end{figure}

SIHA resolves this dependency through structure-free initialization followed by structure-guided refinement. As illustrated in Figure~\ref{fig:framework}, the structure-agnostic hidden-state predictor (HSP-sa) first reconstructs the hidden trajectories from the visible observations alone. This initial estimate completes the multi-agent trajectory set and enables the pretrained NRI backbone~\cite{kipf2018neural} to produce a provisional complete interaction structure and future-state prediction.

Although this initialization enables the first structural estimate, the HSP-sa reconstruction itself does not exploit relational information. SIHA therefore feeds the predicted structure back into the structure-guided hidden-state predictor (HSP-sg), which refines the hidden trajectories under structural guidance. The refined trajectories are then passed to NRI again to update the interaction structure and future prediction. Repeating this process forms an iterative state--structure refinement loop, in which reconstructed states support structural inference and the inferred structure in turn provides additional information for hidden-state reconstruction.

The refinement loop is implemented differently during training and deployment. During HSP-sg training, SIHA maintains a structure cache for each training sample so that structural guidance remains stable across optimization steps. The cache is initialized by the pretrained HSP-sa--NRI pipeline, kept fixed during an initial warm-up period, and then refreshed periodically using the current HSP-sg reconstruction and the pretrained NRI model. This delayed and periodic update reduces the influence of unreliable hidden state estimates at the early stage of training and stabilizes structure-guided reconstruction. At deployment, all model parameters are fixed. The structure is first initialized by HSP-sa and NRI, after which HSP-sg and NRI are directly alternated for a fixed number of refinement rounds to successively update the hidden trajectories and interaction structure.

In the current implementation, NRI serves as both the structure-inference and future-prediction backbone. We use two NRI edge types, whose semantics depend on the underlying system, such as the absence or presence of an interaction in Springs and repulsive or attractive interactions in Charged Particles. We denote the resulting interaction estimate by
$\hat{\mathbf{A}}\in\mathbb{R}^{N\times N}$ in the following sections.

\subsection{Structure-Agnostic Hidden-State Reconstruction}
\label{sec:hsp_sa}

HSP-sa initializes the coupled procedure by estimating the hidden histories from the visible-agent set alone:
\begin{equation}
\text{HSP-sa:}\quad
\mathbf{x}_{\text{vis}}
\mapsto
\hat{\mathbf{x}}_{\text{hid}}^{(0)}
\in\mathbb{R}^{N_{\text{hid}}\times T\times d}.
\end{equation}
It is implemented with a Set Transformer~\cite{lee2019set}. The visible trajectories are treated as an unordered input set, and $N_{\text{hid}}$ learned seed vectors produce an unordered set of hidden-trajectory slots. For the HSP-sa mapping, each visible trajectory is flattened to form $\tilde{\mathbf{x}}_{\text{vis}}\in\mathbb{R}^{N_{\text{vis}}\times(T\cdot d)}$, and the encoder produces
\begin{equation}
\mathbf{Z}=\mathrm{SAB}\!\left(\mathrm{SAB}(\tilde{\mathbf{x}}_{\text{vis}})\right)
\in\mathbb{R}^{N_{\text{vis}}\times D}.
\end{equation}
Here $D$ denotes the latent feature dimension. The HSP-sa decoder maps these visible-agent features to the hidden slots as
\begin{equation}
\hat{\mathbf{X}}_{\text{hid}}
=\mathrm{rFF}\!\left(\mathrm{SAB}\!\left(\mathrm{PMA}_{N_{\text{hid}}}(\mathbf{Z})\right)\right)
\in\mathbb{R}^{N_{\text{hid}}\times(T\cdot d)},
\end{equation}
where unflattening $\hat{\mathbf{X}}_{\text{hid}}$ yields $\hat{\mathbf{x}}_{\text{hid}}^{(0)}\in\mathbb{R}^{N_{\text{hid}}\times T\times d}$. For the synthetic systems, this set-to-set construction is compatible with the permutation ambiguity defined in Section~\ref{sec:problem_formulation}, and the slots are aligned for the supervised objective and evaluation as described in Section~\ref{sec:learning_inference}. In the motion-capture experiments, each output slot is assigned to a predefined masked joint and retains that fixed ordering.

HSP-sa does not receive an estimated graph. Its output is combined with the visible trajectories to form $\hat{\mathbf{x}}^{(0)}=[\mathbf{x}_{\text{vis}},\hat{\mathbf{x}}_{\text{hid}}^{(0)}]$, which the pretrained NRI module maps to an initial structural estimate $\hat{\mathbf{A}}_0$ and future trajectories. The generic SAB, MAB, and PMA definitions are given in \ref{model_details_set}.

\subsection{Structure-Guided Hidden-State Reconstruction}
HSP-sg has the same set-to-set input and output interface as HSP-sa, but additionally conditions its attention computations on the current predicted structure:
\begin{equation}
\text{HSP-sg:}\quad
(\mathbf{x}_{\text{vis}},\hat{\mathbf{A}})
\mapsto
\hat{\mathbf{x}}_{\text{hid}}.
\end{equation}
\begin{figure}[H]
    \centering
    \includegraphics[width=0.85\linewidth]{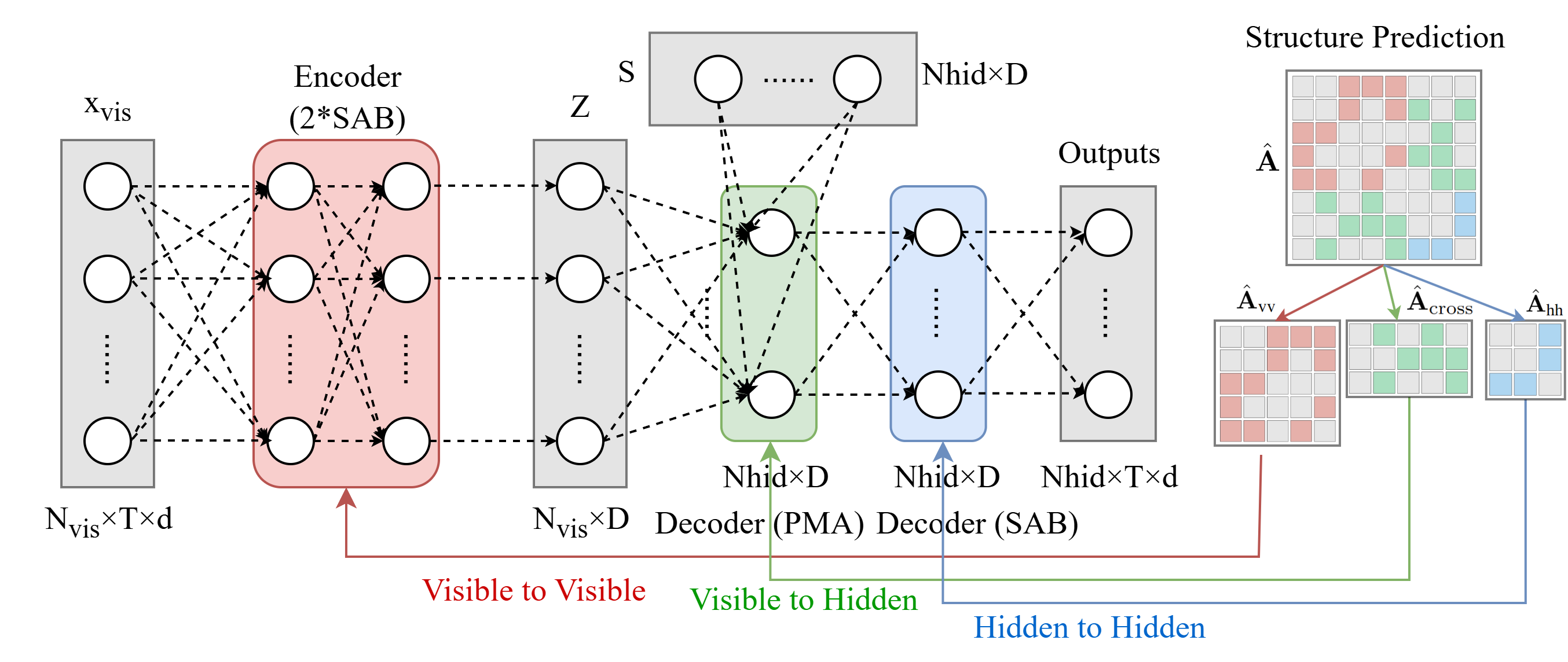}
    \caption{
        Illustration of the structure-guided hidden-state predictor (HSP-sg). The model takes visible trajectories \(\mathbf{x}_{\text{vis}}\) and predicts hidden trajectories with multi-strength structural guidance. The encoder (red) captures visible-to-visible relations; the decoder applies PMA (green) between hidden-slot queries and visible features and SAB (blue) among hidden slots. The structural guidance uses \(\hat{\mathbf{A}}_{vv}\), \(\hat{\mathbf{A}}_{\mathrm{cross}}\), and \(\hat{\mathbf{A}}_{hh}\) in the corresponding attention modules.}

    \label{fig:hsp-sg-architecture}
\end{figure}
As shown in Figure~\ref{fig:hsp-sg-architecture}, the encoder self-attention operates among visible-agent features, PMA connects the hidden output slots with encoded visible features, and the decoder self-attention operates among hidden slots. SIHA uses the corresponding blocks of the predicted adjacency matrix to guide these three attention operations. Specifically,
\[
\hat{\mathbf{A}} =
\begin{bmatrix}
\hat{\mathbf{A}}_{\text{vv}} & \hat{\mathbf{A}}_{\text{vh}} \\
\hat{\mathbf{A}}_{\text{hv}} & \hat{\mathbf{A}}_{\text{hh}}
\end{bmatrix},
\]
where \( \hat{\mathbf{A}}_{\text{vv}} \in \mathbb{R}^{N_{\text{vis}} \times N_{\text{vis}}} \), \( \hat{\mathbf{A}}_{\text{vh}} \in \mathbb{R}^{N_{\text{vis}} \times N_{\text{hid}}} \), \( \hat{\mathbf{A}}_{\text{hv}} \in \mathbb{R}^{N_{\text{hid}} \times N_{\text{vis}}} \), and \( \hat{\mathbf{A}}_{\text{hh}} \in \mathbb{R}^{N_{\text{hid}} \times N_{\text{hid}}} \). Under the convention in Section~\ref{sec:hidden_setting}, $A_{ij}$ continues to represent an interaction from agent $i$ to agent $j$. For the PMA cross-attention, the two cross-agent blocks describe interactions between visible and hidden agents in opposite directions. To use them in the same hidden-query--visible-key attention matrix, we transpose $\hat{\mathbf{A}}_{\text{vh}}$ and average the two estimates:
\[
\hat{\mathbf{A}}_{\mathrm{cross}}
=
\frac{1}{2}
\left(
\hat{\mathbf{A}}_{\text{hv}}
+
\hat{\mathbf{A}}_{\text{vh}}^{\top}
\right)
\in
\mathbb{R}^{N_{\text{hid}}\times N_{\text{vis}}}.
\]
This cross-set matrix is then used to guide the PMA attention between hidden slots and visible-agent features.

HSP-sg applies these structural blocks to the corresponding attention operations as follows:
\begin{itemize}
  \item \( \hat{\mathbf{A}}_{\text{vv}} \) guides the encoder self-attention among visible agents;
  \item \( \hat{\mathbf{A}}_{\mathrm{cross}} \) guides the PMA cross-attention between hidden-slot queries and encoded visible features; and
  \item \( \hat{\mathbf{A}}_{\text{hh}} \) guides the decoder self-attention among hidden slots.
\end{itemize}

For the encoder self-attention, for example, head $i$ receives a structural bias scaled by $\alpha_i$:
\begin{equation}
\begin{aligned}
\text{head}^{\text{sg}}_i &= \text{softmax} \bigg( 
 \frac{\mathbf{Q} \mathbf{W}^Q_i (\mathbf{K} \mathbf{W}^K_i)^\top}{\sqrt{d/h}} 
- \alpha_i \big( \mathbf{1}_{N_{\text{vis}} \times N_{\text{vis}}} - \hat{\mathbf{A}}_{\text{vv}} \big) 
\bigg) 
\mathbf{V} \mathbf{W}^V_i
\end{aligned}
\end{equation}

The term $\mathbf{1}_{N_{\text{vis}}\times N_{\text{vis}}}-\hat{\mathbf{A}}_{\text{vv}}$ downweights attention between pairs not connected in the current estimate. The four attention heads use $\alpha_i\in\{0,1,5,10^9\}$: the first remains unguided, the middle two use soft structural biases, and the last approximates a hard mask. This multi-strength design retains an unguided attention path while exposing other heads to different strengths of the predicted structure. The same construction is applied to PMA cross-attention with $\hat{\mathbf{A}}_{\mathrm{cross}}$ and hidden-slot self-attention with $\hat{\mathbf{A}}_{\text{hh}}$; generic Set Transformer equations are provided in \ref{model_details_set}.

\subsection{Iterative State--Structure Refinement}

Let $f_{\text{pre}}$, $f_{\text{sg}}$, and $g_{\text{NRI}}$ denote HSP-sa, HSP-sg, and the structure estimator of the pretrained NRI module, respectively. With $r$ indexing cache refreshes during training or refinement rounds during deployment, the state--structure updates are
\begin{equation}
\begin{aligned}
\hat{\mathbf{x}}_{\text{hid}}^{(0)}
&=f_{\text{pre}}(\mathbf{x}_{\text{vis}}),\\
\hat{\mathbf{A}}^{(0)}=\mathbf{A}_{\text{cache}}^{(0)}
&=g_{\text{NRI}}\!\left([\mathbf{x}_{\text{vis}},\hat{\mathbf{x}}_{\text{hid}}^{(0)}]\right),\\
\hat{\mathbf{x}}_{\text{hid}}^{(r+1)}
&=f_{\text{sg}}\!\left(\mathbf{x}_{\text{vis}},\mathbf{A}_{\text{cache}}^{(r)}\right),\\
\hat{\mathbf{A}}^{(r+1)}
&=g_{\text{NRI}}\!\left([\mathbf{x}_{\text{vis}},\hat{\mathbf{x}}_{\text{hid}}^{(r+1)}]\right),
\qquad
\mathbf{A}_{\text{cache}}^{(r+1)}\leftarrow\hat{\mathbf{A}}^{(r+1)}.
\end{aligned}
\end{equation}
Thus each refinement first updates the hidden trajectories under the current cached structure and then recomputes the structure from the visible and reconstructed trajectories.

During HSP-sg training, a recurrence step is applied only at a scheduled cache refresh: the cache remains fixed during an initial warm-up period and between refreshes. Separate caches are maintained for the training, validation, and test samples. At deployment, the test cache is initialized by the HSP-sa--NRI pass and one recurrence step is applied in each of a fixed number of rounds. Thus the iterative path changes the reconstructed trajectories and cached structures while using the trained modules with fixed parameters at deployment. The cache schedule and number of deployment rounds are reported in \ref{training_details}.

\subsection{Learning and Inference Procedures}
\label{sec:learning_inference}

SIHA does not introduce a single joint loss, NRI and the two hidden-state predictors retain their respective objectives. The NRI module is pretrained on complete trajectories using the standard NRI evidence lower bound~\cite{kipf2018neural}, whose trajectory-prediction term supports learning the latent graph without ground-truth edge labels. For HSP-sa and HSP-sg, the training target is the ground-truth trajectory of the masked agents. In the synthetic systems, the predicted hidden slots are unordered, so Hungarian matching~\cite{kuhn1955hungarian} aligns them with the target trajectories before mean squared error is evaluated. In the motion-capture experiments, the masked joints have predefined semantic identities, so direct joint-wise MSE is evaluated in their fixed order. The dataset-dependent loss is
\begin{equation}
\mathcal{L}_{\text{HSP-sa}} = \mathcal{L}_{\text{HSP-sg}} =
\begin{cases}
\mathrm{MSE}\!\left(
\mathrm{Align}(\hat{\mathbf{x}}_{\text{hid}},\mathbf{x}_{\text{hid}}),
\mathbf{x}_{\text{hid}}
\right), & \text{synthetic systems},\\
\mathrm{MSE}\!\left(
\hat{\mathbf{x}}_{\text{hid}},
\mathbf{x}_{\text{hid}}
\right), & \text{motion capture}.
\end{cases}
\end{equation}

Training proceeds in two stages. First, HSP-sa and NRI are pretrained separately: HSP-sa receives visible trajectories as input and hidden trajectories as supervised targets, whereas NRI receives complete trajectories and is optimized with the standard NRI objective. Second, HSP-sa initializes the structure cache and HSP-sg is optimized with the dataset-appropriate hidden-state loss above. The pretrained NRI module processes $[\mathbf{x}_{\text{vis}},\hat{\mathbf{x}}_{\text{hid}}]$ to refresh the cache at the prescribed intervals. Detailed settings and pseudocode are given in \ref{training_details}.

At deployment, HSP-sa first reconstructs the hidden trajectories from $\mathbf{x}_{\text{vis}}$, and NRI uses the resulting completed trajectories to initialize the structure and future prediction. HSP-sg and NRI then alternate for the fixed refinement rounds described above. The last HSP-sg output supplies the hidden histories, and the final NRI pass supplies the complete interaction estimate and future trajectories.

\section{Experiments}
\label{experiments}
\subsection{Experimental Setup}
\label{sec:experimental_setup}

\paragraph{Datasets.}
The synthetic evaluation uses the Springs, Charged Particles, and Kuramoto systems, following the trajectory-based structural-inference setting of NRI~\cite{kipf2018neural}. The number of visible agents is fixed at $N_{\text{vis}}=5$, and the reported fixed-count settings use $N_{\text{hid}}\in\{1,2,3,4,5\}$. Each trajectory contains $50$ time steps for training and validation and $100$ time steps for testing. Springs and Charged Particles use four-dimensional position--velocity states, whereas Kuramoto uses the three-dimensional representation specified in \ref{training_details}. Models are configured for the corresponding value of $N_{\text{hid}}$ in these fixed-count evaluations. Section~\ref{sec:mocap} additionally considers recorded human-motion trajectories from the CMU Motion Capture database~\cite{CMUMocap2003}, with one complete limb artificially masked. Dataset sizes, system-specific settings, and hyperparameters remain in \ref{training_details}.

\paragraph{Methods and comparison protocol.}
The experiments instantiate the training and deployment setting defined in Section~\ref{sec:problem_formulation}. HSP-sa and HSP-sg are trained with ground-truth trajectories of masked agents as hidden-state reconstruction targets, while interaction labels are not used for training. HSP-sg, together with the iterative state--structure refinement procedure, constitutes the complete method reported as SIHA (HSP-sg). HSP-sa is the internal controlled baseline: it uses the same hidden-state supervision but omits predicted-structure guidance and iterative refinement. For its structure and future-state outputs, the HSP-sa reconstruction is followed by the same pretrained NRI module.

Standard NRI~\cite{kipf2018neural} is the principal external reference. In this comparison it is trained and evaluated on the visible trajectories only. NRI and SIHA can therefore be compared numerically on visible-agent future prediction and visible-to-visible structure inference, which are defined for both methods. NRI does not produce hidden histories, hidden-agent future trajectories, or hidden-related edges under this protocol, so the corresponding entries are undefined. It therefore serves as an external reference, not a supervision-matched hidden-agent baseline.

\paragraph{Metrics and reporting.}
Hidden-state reconstruction is measured by $\mathrm{MSE}_{\mathrm{HSP}}$. Future-state prediction is measured separately for visible and hidden agents by $\mathrm{MSE}_{\mathrm{FSP,Vis.}}$ and $\mathrm{MSE}_{\mathrm{FSP,Hid.}}$. Structural accuracy is decomposed into visible-to-visible, visible-to-hidden, and hidden-to-hidden blocks, denoted by $\mathrm{ACC}_{\mathrm{V\mbox{-}V}}$, $\mathrm{ACC}_{\mathrm{V\mbox{-}H}}$, and $\mathrm{ACC}_{\mathrm{H\mbox{-}H}}$, respectively. MSE is lower-is-better, whereas structural accuracy is higher-is-better. Ground-truth edges are used only to compute these offline structural metrics. Table~\ref{tab:main_results} reports point estimates over the full experimental grid, while Appendix~C.3 provides five-run mean$\pm$standard-deviation results for three representative settings with $N_{\text{hid}}=3$.

\subsection{Overall Performance}
\label{sec:overall_performance}

Table~\ref{tab:main_results} provides the broad synthetic comparison across all three systems and all five fixed hidden-agent counts. Each table entry lists NRI, HSP-sa, and SIHA (HSP-sg) in that order.
\begin{table*}[t]
\centering
\caption{Main results on three datasets with varying number of hidden agents (\(N_\text{hid}=1,\dots,5\)). 
Metrics: hidden-state prediction MSE, future-state prediction MSE, and structure prediction ACC. Each item reports \textbf{NRI / HSP-sa / SIHA (HSP-sg)}; dashes mark outputs that are not defined for visible-only NRI. Best values among methods for which a metric is defined are bolded.}
\resizebox{\textwidth}{!}{
\begin{tabular}{ccccccccccccc}
\toprule
\multirow{2}{*}{Dataset} & \multirow{2}{*}{\(N_\text{hid}\)} & 
\multicolumn{1}{c}{$\text{MSE}_{\text{HSP}} \,(\downarrow)$} & 
\multicolumn{2}{c}{$\text{MSE}_{\text{FSP}} \,(\downarrow)$} & 
\multicolumn{3}{c}{$\text{ACC} \,(\uparrow)$} \\ 

\cmidrule(lr){3-3} \cmidrule(lr){4-5} \cmidrule(lr){6-8}
 &  &  & Vis. & Hid. & \(\text{V-V}\) & \(\text{V-H}\) & \(\text{H-H}\) \\ 
\midrule
 & 1 & - / 2.5e-3 / \textbf{2.1e-3} & 2.1e-5 / 1.5e-5 / \textbf{1.4e-5} & - / 1.7e-2 / \textbf{1.1e-2} & 89.7 / \textbf{99.7} / \textbf{99.7} & - / 98.2 / \textbf{98.4} & - / - / - \\
 
& 2 & - / 5.6e-3 / \textbf{3.2e-3} & 3.0e-5 / 9.3e-6 / \textbf{7.6e-6} & - / 1.6e-2 / \textbf{9.4e-3} & 76.1 / 98.7 / \textbf{98.9} & - / 91.4 / \textbf{95.8} & - / 68.1 / \textbf{79.0} \\

Springs & 3 & - / 7.1e-3 / \textbf{4.9e-3} & 3.7e-5 / 6.4e-6 / \textbf{5.0e-6} & - / 1.3e-2 / \textbf{1.0e-2} & 72.9 / 97.0 / \textbf{97.2} & - / 83.5 / \textbf{88.6} & - / 61.6 / \textbf{67.4} \\

& 4 & - / 8.1e-3 / \textbf{7.1e-3} & 3.9e-5 / 5.7e-6 / \textbf{5.0e-6} & - / 1.2e-2 / \textbf{1.1e-2} & 70.3 / 95.6 / \textbf{96.0} & - / 78.6 / \textbf{81.7} & - / 57.9 / \textbf{60.5} \\

& 5 & - / 9.1e-3 / \textbf{8.8e-3} & 4.3e-5 / \textbf{6.0e-6} / \textbf{6.0e-6} & - / 1.2e-2 / \textbf{1.1e-2} & 68.8 / 93.0 / \textbf{93.1} & - / 75.2 / \textbf{75.9} & - / 56.3 / \textbf{57.0} \\

\midrule
 & 1 & - / 2.1e-2 / \textbf{1.9e-2} & \textbf{1.4e-3} / \textbf{1.4e-3} / \textbf{1.4e-3} & - / 2.0e-1 / \textbf{1.8e-1} & 71.4 / \textbf{79.5} / \textbf{79.5} & - / 70.6 / \textbf{71.1} & - / - / - \\
 & 2 & - / 3.8e-2 / \textbf{3.7e-2} & \textbf{2.0e-3} / 2.1e-3 / 2.1e-3 & - / 1.7e-1 / \textbf{1.6e-1} & 66.8 / 74.5 / \textbf{74.9} & - / 60.4 / \textbf{60.7} & - / 52.4 / \textbf{53.2} \\
Charged & 3 & - / 4.3e-2 / \textbf{4.1e-2} & \textbf{2.5e-3} / 2.6e-3 / 2.6e-3 & - / 1.4e-1 / \textbf{1.2e-1} & 62.1 / 71.2 / \textbf{72.3} & - / 57.8 / \textbf{58.5} & - / 51.5 / \textbf{51.8} \\
 & 4 & - / 4.5e-2 / \textbf{4.4e-2} & \textbf{2.2e-3} / 2.4e-3 / 2.4e-3 & - / 1.3e-1 / \textbf{1.2e-1} & 61.4 / 68.7 / \textbf{69.6} & - / 56.2 / \textbf{57.4}  & - / 51.6 / \textbf{52.6} \\
 & 5 & - / 4.6e-2 / \textbf{4.5e-2} & \textbf{3.3e-3} / 3.6e-3 / 3.6e-3 & - / \textbf{1.1e-1} / \textbf{1.1e-1} & 59.0 / \textbf{66.6} / \textbf{66.6} & - / 54.5 / \textbf{54.7} & - / \textbf{51.7} / \textbf{51.7} \\

 \midrule
 & 1 & - / 3.6e-2 / \textbf{2.8e-2} & 4.3e-2 / \textbf{1.3e-2} / \textbf{1.3e-2} & - / 4.7e-2 / \textbf{4.0e-2} & 85.0 / 91.7 / \textbf{91.8} & - / 87.6 / \textbf{88.6} & - / - / - \\
 & 2 & - / 1.0e-1 / \textbf{9.5e-2} & 5.0e-2 / 1.5e-2 / \textbf{1.4e-2} & - / 1.1e-1 / \textbf{9.6e-2} & 72.3 / 86.4 / \textbf{86.7} & - / 73.7 / \textbf{74.7} & - / 54.6 / \textbf{55.3} \\
Kuramoto & 3 & - / 1.1e-1 / \textbf{1.0e-1} & 4.2e-2 / 1.6e-2 / \textbf{1.5e-2} & - / 1.3e-1 / \textbf{1.2e-1} & 65.2 / 83.8 / \textbf{84.0} & - / 68.8 / \textbf{70.0} & - / 53.5 / \textbf{54.5} \\
 & 4 & - / 1.2e-1 / \textbf{1.0e-1} & 4.0e-2 / \textbf{1.5e-2} / \textbf{1.5e-2} & - / 1.3e-1 / \textbf{1.2e-1} & 58.4 / 75.4 / \textbf{75.7} & - / \textbf{62.4} / \textbf{62.4}  & - / 51.8 / \textbf{52.0} \\
 & 5 & - / 1.2e-1 / \textbf{1.1e-1} & 3.7e-2 / \textbf{1.6e-2} / \textbf{1.6e-2} & - / \textbf{1.2e-1} / \textbf{1.2e-1} & 58.3 / \textbf{68.5} / \textbf{68.5} & - / \textbf{58.6} / \textbf{58.6} & - / 51.5 / \textbf{51.6} \\
\bottomrule
\end{tabular}}

\label{tab:main_results}
\end{table*}
For the external NRI--SIHA comparison, performance on the common metrics varies across datasets. Both HSP-based configurations achieve higher $\mathrm{ACC}_{\mathrm{V\mbox{-}V}}$ than visible-only NRI throughout the reported grid. Their visible-agent forecasting errors are lower on Springs and Kuramoto, while NRI is comparable or slightly better on Charged. The remaining hidden-related metrics highlight an additional capability of the HSP-based pipelines, since visible-only NRI does not produce these outputs.

The internal comparison between HSP-sa and SIHA (HSP-sg) holds hidden-state supervision and the NRI module fixed while changing structural guidance and refinement. Across the reported point estimates, SIHA (HSP-sg) is better than or equal to HSP-sa on the displayed metrics, with the largest differences generally appearing on Springs and smaller differences on Charged and Kuramoto. We conjecture that this difference may be related to the quality of the structures inferred by the underlying NRI model: its structural inference accuracy is lower on Charged and Kuramoto than on Springs, which may limit the benefit of the NRI-guided iterative refinement in HSP-sg. Since the SIHA refinement mechanism only requires a structural estimate together with future-state prediction, it could in principle be paired with other compatible structural-inference backbones. We therefore expect that a stronger backbone may further improve the effectiveness of HSP-sg, although this remains to be verified experimentally.

\subsection{Effect of the Number of Hidden Agents}
\label{sec:hidden_count_effect}

Table~\ref{tab:main_results} shows a clear degradation in performance as the number of hidden agents increases while $N_{\text{vis}}=5$ remains fixed. As the latent portion of the system grows while the observed set remains unchanged, hidden-state reconstruction becomes progressively more difficult, and structural inference accuracy also generally declines. This trend is observed across Springs, Charged, and Kuramoto. Despite this increasing difficulty, the HSP-based pipelines consistently maintain higher visible-to-visible structural accuracy than the visible-only NRI reference across the reported hidden-agent counts. Additionally, the models in these experiments assume a fixed hidden-agent count; the setting in which this count is unknown is examined separately in~\ref{d1}.

\subsection{Motion-Capture Case Study}
\label{sec:mocap}

We evaluate SIHA on walking sequences from subjects \#35 and \#69 of the CMU Motion Capture database~\cite{CMUMocap2003}. Each frame contains 31 body joints, each represented by 3D position and 3D velocity. We simulate two whole-limb occlusion settings. The left-arm setting masks seven joints---\texttt{lclavicle}, \texttt{lhumerus}, \texttt{lradius}, \texttt{lwrist}, \texttt{lhand}, \texttt{lfingers}, and \texttt{lthumb}---leaving 24 visible joints. The left-leg setting masks five joints---\texttt{lhipjoint}, \texttt{lfemur}, \texttt{ltibia}, \texttt{lfoot}, and \texttt{ltoes}---leaving 26 visible joints. These experiments apply artificial whole-limb masks to recorded human-motion trajectories and are not a benchmark of occlusion caused by a particular camera or sensor.

HSP-sa and HSP-sg are trained on fully observed motion-capture sequences, with the ground-truth trajectories of the designated masked joints used only as training-time hidden-state reconstruction targets; deployment and evaluation receive only the visible joints. Training and evaluation use direct joint-wise MSE in the predefined masked-joint order. For each subject, SIHA uses an NRI backbone pretrained on that subject's full 31-joint sequences for structure inference and future prediction. The same subject-specific full-observation NRI is reused for the left-arm and left-leg experiments. The external visible-only NRI baseline is trained and evaluated on 24 visible joints for the left-arm setting and 26 visible joints for the left-leg setting.

\begin{table}[h]
\centering
\caption{Results on two CMU motion-capture subjects under two whole-limb occlusion settings.}
\label{tab:mocap}
\small
\setlength{\tabcolsep}{4pt}
\begin{tabular}{lllccc}
\toprule
Subject & Occlusion & Method & $\mathrm{MSE}_{\mathrm{HSP}} \downarrow$ & $\mathrm{MSE}_{\mathrm{FSP,Hid.}} \downarrow$ & $\mathrm{MSE}_{\mathrm{FSP,Vis.}} \downarrow$ \\
\midrule
\multirow{6}{*}{CMU \#35}
& \multirow{3}{*}{Left arm}
& NRI & --- & --- & 0.012463 \\
& & HSP-sa & 0.015547 & 0.039196 & 0.008176 \\
& & SIHA (HSP-sg) & \textbf{0.008056} & \textbf{0.021843} & \textbf{0.007788} \\
\cmidrule(lr){2-6}
& \multirow{3}{*}{Left leg}
& NRI & --- & --- & 0.039669 \\
& & HSP-sa & 0.025210 & 0.022922 & 0.006773 \\
& & SIHA (HSP-sg) & \textbf{0.023735} & \textbf{0.014077} & \textbf{0.006594} \\
\midrule
\multirow{6}{*}{CMU \#69}
& \multirow{3}{*}{Left arm}
& NRI & --- & --- & 0.001954 \\
& & HSP-sa & \textbf{0.003224} & 0.004434 & 0.001304 \\
& & SIHA (HSP-sg) & 0.003349 & \textbf{0.004263} & \textbf{0.001227} \\
\cmidrule(lr){2-6}
& \multirow{3}{*}{Left leg}
& NRI & --- & --- & 0.001635 \\
& & HSP-sa & 0.001880 & 0.003977 & 0.000939 \\
& & SIHA (HSP-sg) & \textbf{0.001616} & \textbf{0.003633} & \textbf{0.000915} \\
\bottomrule
\end{tabular}
\end{table}

Across the four subject--occlusion settings, SIHA (HSP-sg) improves both hidden- and visible-agent future prediction over HSP-sa. It also reduces hidden-history reconstruction error in three settings; the exception is the subject \#69 left-arm case. On the common visible-future metric, both HSP-based configurations have lower errors than the corresponding visible-only NRI baseline in all four settings. Overall, the expanded results show that structure-guided refinement provides consistent future-prediction benefits across the evaluated subjects and limb-occlusion patterns while generally improving hidden-history reconstruction.

\begin{figure}[h]
    \centering
    \subfloat[\label{fig:full_nri_dual}]{%
        \includegraphics[width=0.11\textwidth]{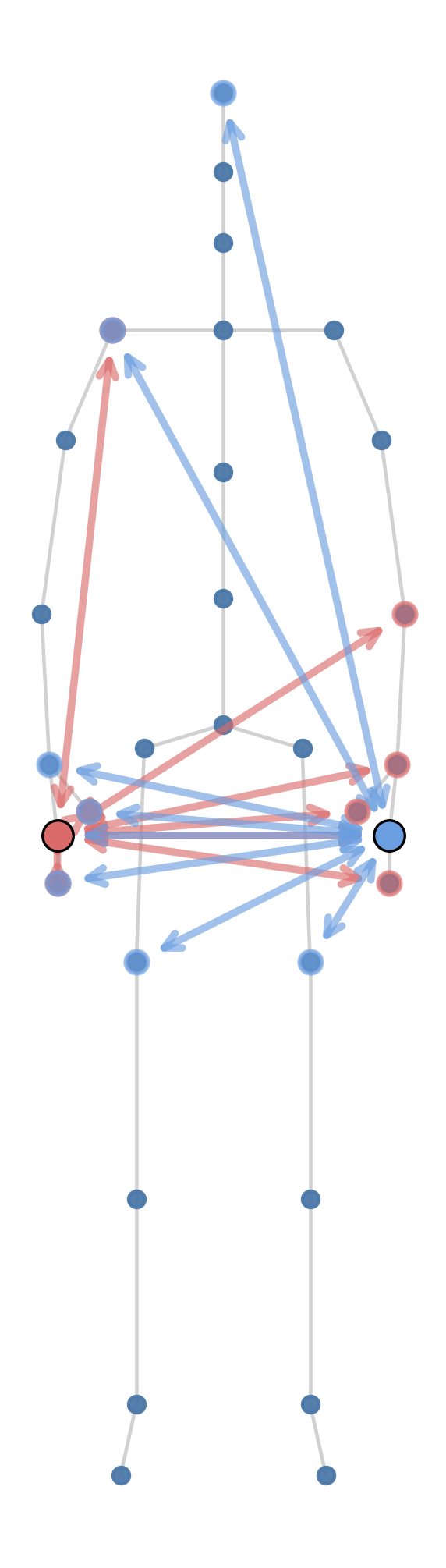}}
    \hfill
    \subfloat[\label{fig:visible_only_nri}]{%
        \includegraphics[width=0.09\textwidth]{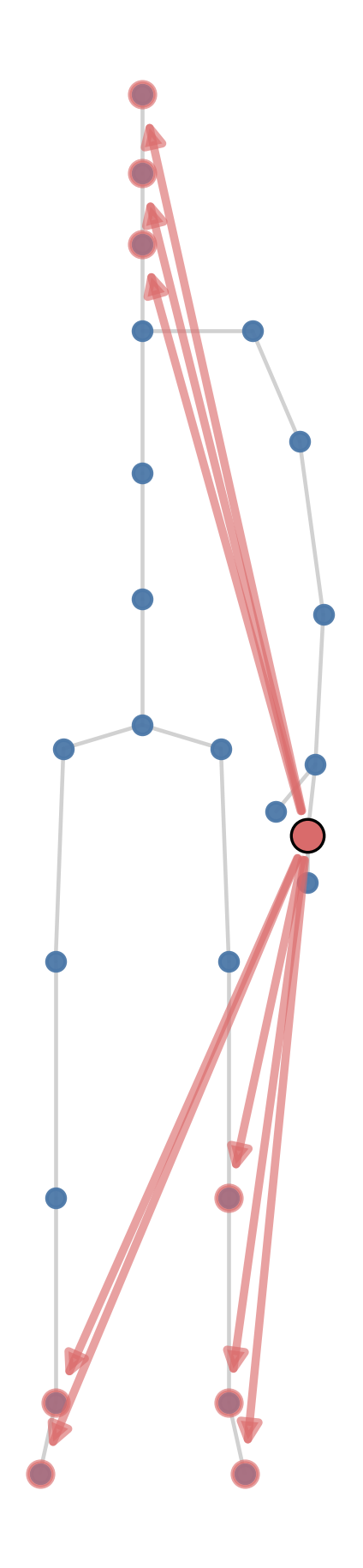}}
    \hfill
    \subfloat[\label{fig:hsp_sa_left}]{%
        \includegraphics[width=0.11\textwidth]{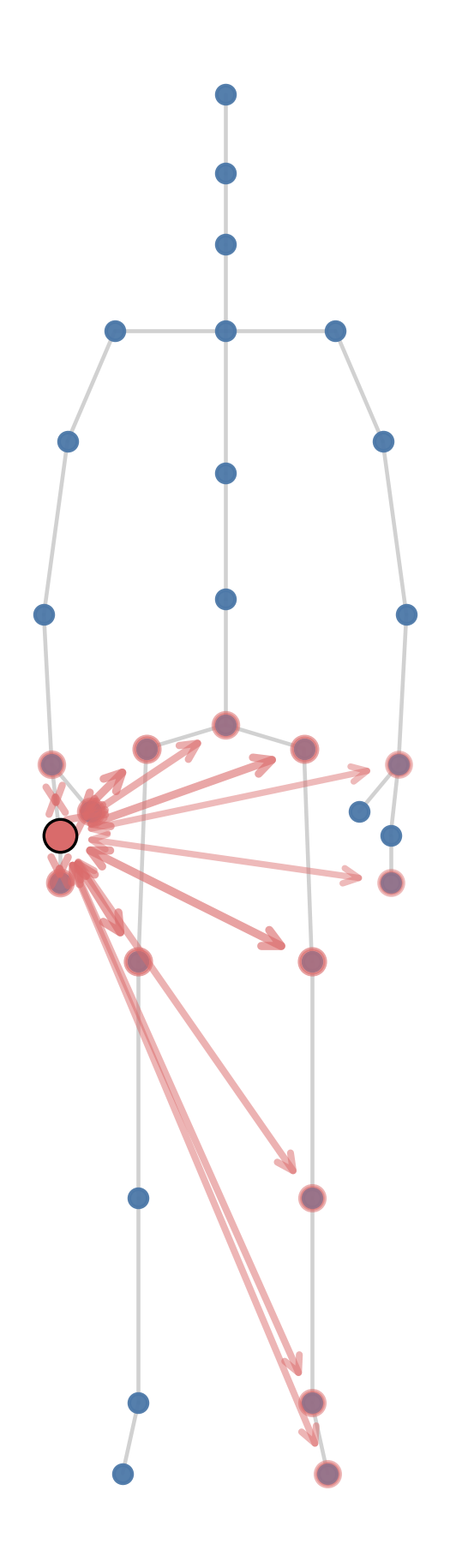}}
    \hfill
    \subfloat[\label{fig:hsp_sa_right}]{%
        \includegraphics[width=0.11\textwidth]{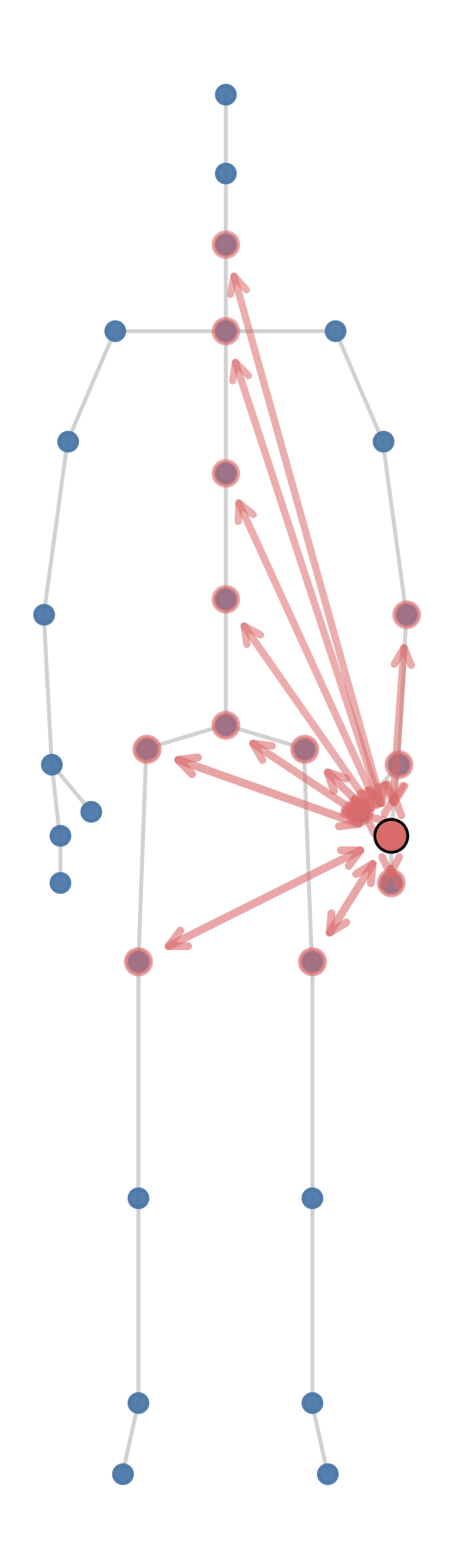}}
    \hfill
    \subfloat[\label{fig:hsp_sg_left}]{%
        \includegraphics[width=0.11\textwidth]{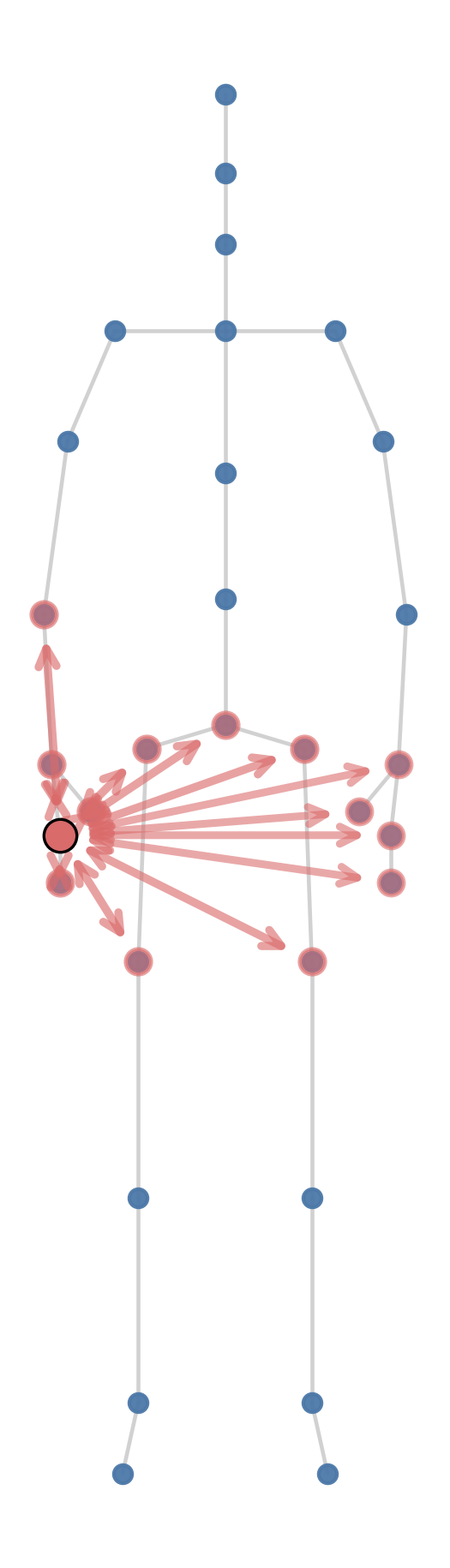}}
    \hfill
    \subfloat[\label{fig:hsp_sg_right}]{%
        \includegraphics[width=0.11\textwidth]{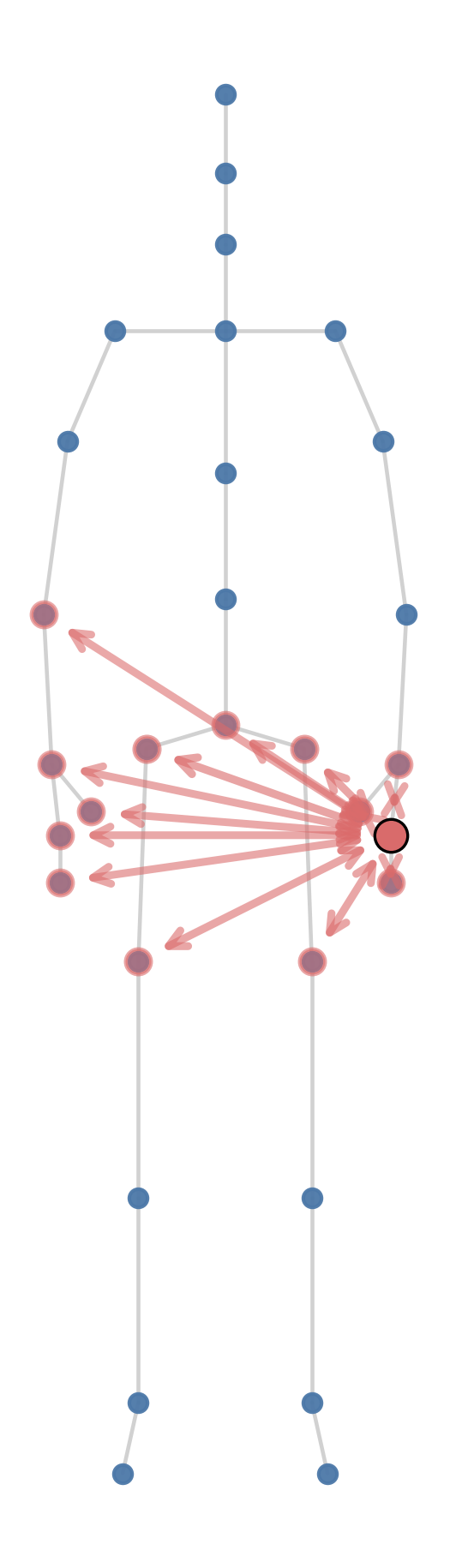}}

    \caption{Visualizations of the left/right hand focus on the motion-capture limb-occlusion experiment. (a) Full-observation NRI reference; (b) visible-only NRI; (c) HSP-sa with left-hand focus; (d) HSP-sa with right-hand focus; (e) SIHA with left-hand focus; (f) SIHA with right-hand focus.}
    \label{fig:mocap_six_panel}
\end{figure}

For qualitative visualization, we retain the subject \#35 left-arm setting in Figure~\ref{fig:mocap_six_panel}, with comparisons focused on the left and right hands. Human walking naturally involves coordinated motion between the two arms, so a strong relation between the left and right hands is expected. Consistent with this intuition, the NRI model trained on complete trajectories assigns the strongest connections of the right-hand node mainly to joints in the left-hand region. The visible-only NRI baseline has no hidden-arm nodes in its input and therefore cannot recover these cross-limb relations. After hidden-state reconstruction, both HSP-based pipelines can infer such connections, while HSP-sg produces a more concentrated cross-hand pattern than HSP-sa and is therefore closer to the full-observation NRI reference. Together with the quantitative improvements in Table~\ref{tab:mocap}, this qualitative result suggests that HSP-sg achieves a stronger coupling between hidden-state prediction and structural inference. The focus-score definition, edge-selection thresholds, and panel-specific settings are given in~\ref{app:mocap_details}.

\subsection{Model and Supervision Analysis}
\label{sec:model_supervision}

\paragraph{Multi-strength structural guidance.}
We analyze the attention guidance mechanism on Springs with \(N_{\text{hid}}=3\). Three HSP-sg variants modify the attention-head coefficients to \([1,1,5,1e9]\) (no unguided head), \([0,1,5,5]\) (no hard-mask head), and \([0,0,1e9,1e9]\) (no intermediate-strength guidance).

Table~\ref{tab:ablation} shows that removing the unguided head causes the largest degradation across the reported metrics, indicating that retaining a fully learnable attention path is important. Removing the intermediate-strength guidance also reduces performance, although to a smaller extent. In contrast, removing the hard-mask head leaves the reported point estimates unchanged in this setting, suggesting that the main benefit of the multi-strength design comes from combining unguided and softly structure-guided attention. Additionally, the strong masking head may also play a role when accurate structural information is available.

\begin{table}[h]
\centering
\caption{Ablation study on the Springs dataset with $N_{\text{hid}}=3$.}
\small
\setlength{\tabcolsep}{6pt}
\begin{tabular}{lccc}
\toprule
Model
& $\mathrm{MSE}_{\mathrm{HSP}}$
& $\mathrm{MSE}_{\mathrm{FSP,Vis.}}$
& $\mathrm{ACC}_{\mathrm{V\mbox{-}V}}$ \\
\midrule
SIHA (HSP-sg)       & \textbf{4.9e-3} & \textbf{5.0e-6} & \textbf{97.2} \\
No Unguided         & 2.0e-2          & 1.3e-5          & 91.5 \\
No Hard Mask        & \textbf{4.9e-3} & \textbf{5.0e-6} & \textbf{97.2} \\
No Intermediate     & 8.2e-3          & 6.1e-6          & 96.6 \\
\bottomrule
\end{tabular}
\label{tab:ablation}
\end{table}

\paragraph{Dependence on hidden-state supervision.}
We further examine three HSP-sa variants without direct supervision from ground-truth hidden trajectories. These variants differ in the initialization and optimization of NRI, while all of them train HSP-sa only through the visible-agent future-prediction objective. \ref{d3} gives their complete definitions and results on Springs for $N_{\text{hid}}\in\{1,2,3,4,5\}$. All three variants perform substantially worse than the supervised HSP-sa configuration in both hidden-state reconstruction and visible-agent prediction, with corresponding degradation in structural inference. These results indicate that, within the current SIHA framework, simply removing hidden-state supervision and relying on visible-future prediction is not sufficient for effective hidden-agent reconstruction and structural inference. Extending SIHA toward a fully latent training setting without hidden-state supervision therefore represents an
important direction for future work.

\subsection{Additional Analyses}
\label{sec:additional_analyses}

Additional analyses are provided in the appendices: \ref{app:encoder_ablation} reports the comparison of HSP-sa encoder. \ref{app:seed_stats} reports five-run mean$\pm$standard-deviation results for the three representative $N_{\text{hid}}=3$ settings. \ref{d1} evaluates selection among models configured for different hidden-agent counts. \ref{d2} reports the effect of additional unmodeled hidden agents on Springs. Detailed optimization settings and computational costs remain in \ref{training_details}.

\section{Conclusion}
In this paper, we investigated structural inference in multi-agent systems when the trajectories of some agents are unavailable at deployment. We proposed Structural Inference under Hidden Agents (SIHA), which reconstructs hidden-agent trajectories and infers the complete interaction structure through coupled state--structure refinement. SIHA first initializes hidden trajectories without structural guidance, estimates interactions using NRI, and then incorporates the inferred structure into a structure-guided hidden-state predictor with multi-strength attention for iterative refinement. Extensive experiments on three benchmark dynamical systems and recorded motion-capture trajectories demonstrate the effectiveness of the proposed framework. SIHA achieves higher visible-to-visible structural accuracy than visible-only NRI across the evaluated synthetic settings while additionally recovering hidden-agent trajectories and interactions involving hidden agents. Compared with the structure-agnostic variant under the same supervision, structure-guided refinement improves or maintains the reported reconstruction, forecasting, and structural metrics. Experiments with different numbers of hidden agents, mechanism ablations, and simulated limb occlusion further demonstrate the effectiveness of the proposed design.

A limitation of the current formulation is its reliance on complete trajectories during training to supervise hidden-state reconstruction, although ground-truth edge labels are not required for model training or deployment. The tested objectives without direct hidden-state supervision do not recover the performance of the supervised formulation, leaving structural inference with fully latent hidden agents an open problem. Future work will investigate self-supervised hidden-state reconstruction and structural inference when hidden-agent trajectories are unavailable during both training and deployment.

\bibliographystyle{unsrt}
\bibliography{references}

@article{pratapa2020benchmarking,
	title={Benchmarking algorithms for gene regulatory network inference from single-cell transcriptomic data},
	author={Pratapa, Aditya and Jalihal, Amogh P and Law, Jeffrey N and Bharadwaj, Aditya and Murali, TM},
	journal={Nature Methods},
	volume={17},
	number={2},
	pages={147--154},
	year={2020},
	doi={10.1038/s41592-019-0690-6},
	publisher={Nature Publishing Group US New York}
}

@article{centola2010spread,
  title={The Spread of Behavior in an Online Social Network Experiment},
  author={Centola, Damon},
  journal={Science},
  volume={329},
  number={5996},
  pages={1194--1197},
  year={2010},
  doi={10.1126/science.1185231}
}

@article{acemoglu2012network,
  title={The Network Origins of Aggregate Fluctuations},
  author={Acemoglu, Daron and Carvalho, Vasco M. and Ozdaglar, Asuman and Tahbaz-Salehi, Alireza},
  journal={Econometrica},
  volume={80},
  number={5},
  pages={1977--2016},
  year={2012},
  doi={10.3982/ECTA9623}
}

@article{nitzan2017revealing,
  title={Revealing Physical Interaction Networks from Statistics of Collective Dynamics},
  author={Nitzan, Mor and Casadiego, Jose and Timme, Marc},
  journal={Science Advances},
  volume={3},
  number={2},
  pages={e1600396},
  year={2017},
  doi={10.1126/sciadv.1600396},
  url={https://doi.org/10.1126/sciadv.1600396}
}

@inproceedings{battaglia2016interaction,
  title={Interaction Networks for Learning about Objects, Relations and Physics},
  author={Battaglia, Peter W. and Pascanu, Razvan and Lai, Matthew and Rezende, Danilo Jimenez and Kavukcuoglu, Koray},
  booktitle={Advances in Neural Information Processing Systems},
  volume={29},
  pages={4502--4510},
  year={2016},
  url={https://proceedings.neurips.cc/paper/2016/hash/3147da8ab4a0437c15ef51a5cc7f2dc4-Abstract.html}
}

@inproceedings{kipf2018neural,
	title={Neural relational inference for interacting systems},
	author={Kipf, Thomas and Fetaya, Ethan and Wang, Kuan-Chieh and Welling, Max and Zemel, Richard},
	booktitle={International Conference on Machine Learning},
	pages={2688--2697},
	year={2018},
	organization={PMLR}
}

@inproceedings{graber2020dynamic,
  title={Dynamic Neural Relational Inference},
  author={Graber, Colin and Schwing, Alexander G.},
  booktitle={IEEE/CVF Conference on Computer Vision and Pattern Recognition (CVPR)},
  pages={8513--8522},
  year={2020},
  url={https://openaccess.thecvf.com/content_CVPR_2020/html/Graber_Dynamic_Neural_Relational_Inference_CVPR_2020_paper.html}
}

@inproceedings{alet2019neural,
  title={Neural Relational Inference with Fast Modular Meta-Learning},
  author={Alet, Ferran and Weng, Erica and Lozano-P{\'e}rez, Tom{\'a}s and Kaelbling, Leslie Pack},
  booktitle={Advances in Neural Information Processing Systems},
  volume={32},
  pages={11804--11815},
  year={2019},
  url={https://proceedings.neurips.cc/paper/2019/hash/b294504229c668e750dfcc4ea9617f0a-Abstract.html}
}

@inproceedings{chen2021neural,
  title={Neural Relational Inference with Efficient Message Passing Mechanisms},
  author={Chen, Siyuan and Wang, Jiahai and Li, Guoqing},
  booktitle={{AAAI} Conference on Artificial Intelligence},
  volume={35},
  number={8},
  pages={7055--7063},
  year={2021},
  doi={10.1609/aaai.v35i8.16868},
  url={https://ojs.aaai.org/index.php/AAAI/article/view/16868}
}

@article{wang2022iterative,
	title={Iterative structural inference of directed graphs},
	author={Wang, Aoran and Pang, Jun},
	journal={Advances in Neural Information Processing Systems},
	volume={35},
	pages={8717--8730},
	year={2022}
}

@inproceedings{kingma2013auto,
  title={Auto-Encoding Variational Bayes},
  author={Kingma, Diederik P. and Welling, Max},
  booktitle={International Conference on Learning Representations},
  year={2014},
  url={https://arxiv.org/abs/1312.6114}
}

@article{zheng2026diffusion,
  title={Diffusion Model for Relational Inference in Interacting Systems},
  author={Zheng, Shuhan and Li, Ziqiang and Fujiwara, Kantaro and Tanaka, Gouhei},
  journal={IEEE Transactions on Network Science and Engineering},
  volume={13},
  pages={1990--2003},
  year={2026},
  doi={10.1109/TNSE.2025.3607563},
  url={https://ieeexplore.ieee.org/document/11164166/}
}

@inproceedings{wang2024conjoined,
  title={Structural Inference of Dynamical Systems with Conjoined State Space Models},
  author={Wang, Aoran and Pang, Jun},
  booktitle={Advances in Neural Information Processing Systems},
  volume={37},
  pages={75355--75391},
  year={2024},
  doi={10.52202/079017-2399},
  url={https://proceedings.neurips.cc/paper_files/paper/2024/hash/89c61fce5a8b73871d1c4073f486b134-Abstract-Conference.html}
}

@inproceedings{tashiro2021csdi,
  title={{CSDI}: Conditional Score-Based Diffusion Models for Probabilistic Time Series Imputation},
  author={Tashiro, Yusuke and Song, Jiaming and Song, Yang and Ermon, Stefano},
  booktitle={Advances in Neural Information Processing Systems},
  volume={34},
  pages={24804--24816},
  year={2021},
  url={https://proceedings.neurips.cc/paper/2021/hash/cfe8504bda37b575c70ee1a8276f3486-Abstract.html}
}

@inproceedings{cini2022filling,
  title={Filling the {Gaps}: {Multivariate Time Series Imputation by Graph Neural Networks}},
  author={Cini, Andrea and Marisca, Ivan and Alippi, Cesare},
  booktitle={International Conference on Learning Representations},
  year={2022},
  url={https://openreview.net/forum?id=kOu3-S3wJ7}
}

@inproceedings{maddison2017concrete,
  title={The Concrete Distribution: A Continuous Relaxation of Discrete Random Variables},
  author={Maddison, Chris J. and Mnih, Andriy and Teh, Yee Whye},
  booktitle={International Conference on Learning Representations},
  year={2017}
}

@inproceedings{lee2019set,
  title={Set transformer: A framework for attention-based permutation-invariant neural networks},
  author={Lee, Juho and Lee, Yoonho and Kim, Jungtaek and Kosiorek, Adam and Choi, Seungjin and Teh, Yee Whye},
  booktitle={International Conference on Machine Learning},
  pages={3744--3753},
  year={2019},
  organization={PMLR}
}

@article{locatello2020object,
  title={Object-centric learning with slot attention},
  author={Locatello, Francesco and Weissenborn, Dirk and Unterthiner, Thomas and Mahendran, Aravindh and Heigold, Georg and Uszkoreit, Jakob and Dosovitskiy, Alexey and Kipf, Thomas},
  journal={Advances in Neural Information Processing Systems},
  volume={33},
  pages={11525--11538},
  year={2020}
}

@article{zhang2019general,
  title={A General Deep Learning Framework for Network Reconstruction and Dynamics Learning},
  author={Zhang, Zhang and Zhao, Yi and Liu, Jing and Wang, Shuo and Tao, Ruyi and Xin, Ruyue and Zhang, Jiang},
  journal={Applied Network Science},
  volume={4},
  pages={110},
  year={2019},
  doi={10.1007/s41109-019-0194-4}
}

@inproceedings{li2020evolvegraph,
  title={{EvolveGraph}: Multi-Agent Trajectory Prediction with Dynamic Relational Reasoning},
  author={Li, Jiachen and Yang, Fan and Tomizuka, Masayoshi and Choi, Chiho},
  booktitle={Advances in Neural Information Processing Systems},
  volume={33},
  pages={19783--19794},
  year={2020},
  url={https://proceedings.neurips.cc/paper/2020/hash/e4d8163c7a068b65a64c89bd745ec360-Abstract.html}
}

@article{grossmann2023masked,
  title={Unsupervised Relational Inference Using Masked Reconstruction},
  author={Gro{\ss}mann, Gerrit and Zimmerlin, Julian and Backenk{\"o}hler, Michael and Wolf, Verena},
  journal={Applied Network Science},
  volume={8},
  pages={18},
  year={2023},
  doi={10.1007/s41109-023-00542-x}
}

@article{han2024collective,
  title={Collective Relational Inference for Learning Heterogeneous Interactions},
  author={Han, Zhichao and Fink, Olga and Kammer, David S.},
  journal={Nature Communications},
  volume={15},
  pages={3191},
  year={2024},
  doi={10.1038/s41467-024-47098-7}
}

@inproceedings{franceschi2019learning,
  title={Learning Discrete Structures for Graph Neural Networks},
  author={Franceschi, Luca and Niepert, Mathias and Pontil, Massimiliano and He, Xiao},
  booktitle={International Conference on Machine Learning},
  volume={97},
  pages={1972--1982},
  year={2019},
  url={https://proceedings.mlr.press/v97/franceschi19a.html}
}

@article{cai2024granger,
  title={{Granger} Causal Representation Learning for Groups of Time Series},
  author={Cai, Ruichu and Wu, Yunjin and Huang, Xiaokai and Chen, Wei and Fu, Tom Z. J. and Hao, Zhifeng},
  journal={Science China Information Sciences},
  volume={67},
  number={5},
  pages={152103},
  year={2024},
  doi={10.1007/s11432-021-3724-0}
}

@article{wang2024credible,
  title={A Credible Traffic Prediction Method Based on Self-Supervised Causal Discovery},
  author={Wang, Dan and Liu, Yingjie and Song, Bin},
  journal={Science China Information Sciences},
  volume={67},
  number={5},
  pages={152303},
  year={2024},
  doi={10.1007/s11432-023-3899-1}
}

@article{yang2021hidden,
  title={Hidden Network Generating Rules from Partially Observed Complex Networks},
  author={Yang, Ruochen and Sala, Frederic and Bogdan, Paul},
  journal={Communications Physics},
  volume={4},
  pages={199},
  year={2021},
  doi={10.1038/s42005-021-00701-5}
}

@article{cui2024continuous,
  title={Learning Continuous Network Emerging Dynamics from Scarce Observations via Data-Adaptive Stochastic Processes},
  author={Cui, Jiaxu and Wang, Qipeng and Sun, Bingyi and Liu, Jiming and Yang, Bo},
  journal={Science China Information Sciences},
  volume={67},
  number={12},
  pages={222206},
  year={2024},
  doi={10.1007/s11432-023-4216-y}
}

@inproceedings{cao2018brits,
  title={{BRITS}: Bidirectional Recurrent Imputation for Time Series},
  author={Cao, Wei and Wang, Dong and Li, Jian and Zhou, Hao and Li, Lei and Li, Yitan},
  booktitle={Advances in Neural Information Processing Systems},
  volume={31},
  pages={6776--6786},
  year={2018},
  url={https://proceedings.neurips.cc/paper/2018/hash/734e6bfcd358e25ac1db0a4241b95651-Abstract.html}
}

@inproceedings{fortuin2020gpvae,
  title={{GP-VAE}: Deep Probabilistic Time Series Imputation},
  author={Fortuin, Vincent and Baranchuk, Dmitry and R{\"a}tsch, Gunnar and Mandt, Stephan},
  booktitle={International Conference on Artificial Intelligence and Statistics},
  volume={108},
  pages={1651--1661},
  year={2020},
  url={https://proceedings.mlr.press/v108/fortuin20a.html}
}

@article{du2023saits,
  title={{SAITS}: Self-Attention-Based Imputation for Time Series},
  author={Du, Wenjie and C{\^o}t{\'e}, David and Liu, Yan},
  journal={Expert Systems with Applications},
  volume={219},
  pages={119619},
  year={2023},
  doi={10.1016/j.eswa.2023.119619}
}

@inproceedings{vaswani2017attention,
  title={Attention Is All You Need},
  author={Vaswani, Ashish and Shazeer, Noam and Parmar, Niki and Uszkoreit, Jakob and Jones, Llion and Gomez, Aidan N. and Kaiser, {\L}ukasz and Polosukhin, Illia},
  booktitle={Advances in Neural Information Processing Systems},
  volume={30},
  pages={5998--6008},
  year={2017},
  url={https://proceedings.neurips.cc/paper/2017/hash/3f5ee243547dee91fbd053c1c4a845aa-Abstract.html}
}

@article{kuhn1955hungarian,
  title={The Hungarian Method for the Assignment Problem},
  author={Kuhn, Harold W.},
  journal={Naval Research Logistics Quarterly},
  volume={2},
  number={1--2},
  pages={83--97},
  year={1955},
  doi={10.1002/nav.3800020109}
}

@misc{CMUMocap2003,
  author = {{Carnegie Mellon University}},
  title  = {{Carnegie-Mellon Motion Capture Database}},
  year   = {2003},
  url    = {http://mocap.cs.cmu.edu}
}

\appendix

\section{Architectural Details}

\subsection{Neural Relational Inference Backbone}
\label{model_details_nri}
NRI learns a latent interaction graph from observed trajectories without ground-truth edge labels~\cite{kipf2018neural}. In SIHA, it is the structure-inference and future-prediction backbone that processes visible trajectories together with reconstructed hidden trajectories.

NRI is built on the VAE framework~\cite{kingma2013auto}. It consists of two core components: a GNN-based encoder and a trajectory-prediction decoder. The encoder takes a sequence of features $\mathbf{x} \in \mathbb{R}^{N \times T \times d}$ and encodes it into a distribution over latent edge types, typically modeled as categorical variables with $K$ classes and represented by $\mathbf{z} \in \mathbb{R}^{N \times N\times K}$:
\begin{equation}
	\mathbf{h}=f_{\mathrm{enc}}(\mathbf{x}),\ q_{\phi}(\mathbf{z}|\mathbf{x})  =\mathrm{softmax}(\mathbf{h}).
\end{equation}

Directly sampling the discrete adjacency tensor $\mathbf{z}$ from $q_{\phi}(\mathbf{z}|\mathbf{x})$ will lead to a non-differentiable process. To allow back-propagation for training, NRI adopts the Gumbel-Softmax trick~\cite{maddison2017concrete}. The inferred graph $\mathbf{z}$ is then passed to the GNN-based decoder, which simulates the next-step dynamics using message passing on the sampled interaction graph. The decoder is trained to minimize the reconstruction loss between predicted and true future states, and is expected to model the interactive dynamic patterns of the system:
\begin{equation}\mathbf{z}_{ij}=\mathrm{softmax}\left((\mathbf{h}_{ij}+\mathbf{g})/\tau\right),\end{equation}\begin{equation}
p_\theta(\mathbf{x}|\mathbf{z})=\prod_{t=1}^Tp_\theta(\mathbf{x}^{t+1}|\mathbf{x}^{1:t},\mathbf{z}).\end{equation}

The NRI model is trained by maximizing the evidence lower bound (ELBO):
\begin{equation}\text{ELBO}=\mathbb{E}_{q_\phi(\mathbf{z}|\mathbf{x})}[\log p_\theta(\mathbf{x}|\mathbf{z})]-D_{\mathrm{KL}}(q_\phi(\mathbf{z}|\mathbf{x})||p(\mathbf{z})),\end{equation}
where the former term corresponds to minimizing the state prediction error, while the latter term serves as the latent space regularization to constrain the discrepancy between the posterior distribution $q_\phi(\mathbf{z}|\mathbf{x})$ in the latent space and the prior distribution $p(\mathbf{z})$ (typically assumed to be uniform).

\subsection{Set Transformer and Hidden-State Predictor Architecture}
\label{model_details_set}
The Set Transformer~\cite{lee2019set} is an attention-based neural network architecture designed to model higher-order interactions among set elements via attention mechanisms.

Given an input set $\mathbf{X} = \{\mathbf{x}_1, \dots, \mathbf{x}_n\}$ with $\mathbf{x}_i \in \mathbb{R}^d$, the encoder of the Set Transformer applies stacked Set Attention Blocks (SABs) to produce latent representations $\mathbf{Z} \in \mathbb{R}^{n \times d}$:
\begin{equation}
	\text{Encoder}(\mathbf{X}) =
	\text{SAB}(\text{SAB}(\mathbf{X})).
\end{equation}

Each SAB captures interactions within the set through a \emph{Multihead Attention Block (MAB)}, defined as:
\begin{equation}
	\text{SAB}(\mathbf{X}) := \text{MAB}(\mathbf{X}, \mathbf{X}),
\end{equation}
where $\mathbf{X} \in \mathbb{R}^{n \times d}$ is both the query and the key/value input.

The MAB computes multihead cross attention between a query set $\mathbf{X}\in\mathbb{R}^{n\times d}$ and a key/value set $\mathbf{Y}\in\mathbb{R}^{m\times d}$:
\begin{equation}
	\text{MAB}(\mathbf{X}, \mathbf{Y}) =
	\text{LN}\big( \mathbf{H} + \text{rFF}( \mathbf{H}) \big) , 
\end{equation}
\begin{equation}
    \text{where} \ \  \mathbf{H} =
	\text{LN}\big( \mathbf{X} + \text{Multihead}( \mathbf{X}, \mathbf{Y}, \mathbf{Y}) \big) ,
\end{equation}

where $\text{LN}(\cdot)$ denotes layer normalization and $\text{rFF}(\cdot)$ is a row-wise feedforward network. The multihead attention mechanism $\text{Multihead}(\mathbf{Q}, \mathbf{K}, \mathbf{V})$~\cite{vaswani2017attention} is defined as:
\begin{equation}
	\text{Multihead}(\mathbf{Q}, \mathbf{K}, \mathbf{V}) =
	\text{Concat}\big( \text{head}_1, \dots, \text{head}_h \big) \mathbf{W}^O,
\end{equation}
where each head is computed by scaled dot-product attention:
\begin{equation}
	\text{head}_i = \text{softmax} \left( \frac{\mathbf{Q} \mathbf{W}^Q_i (\mathbf{K} \mathbf{W}^K_i)^\top}{\sqrt{d/h}} \right) \mathbf{V} \mathbf{W}^V_i.
\end{equation}

Here $h$ denotes the number of attention heads, and $\mathbf{W}^Q_i, \mathbf{W}^K_i, \mathbf{W}^V_i, \mathbf{W}^O$ are learnable projection matrices.

For aggregation and output, the decoder uses a Pooling by Multihead Attention (PMA) module, which maps the latent representations to a fixed-size output set using $k$ learnable seed vectors $\mathbf{S} \in \mathbb{R}^{k \times d}$:

\begin{equation}\mathrm{Decoder}(\mathbf{Z})=\mathrm{rFF}(\mathrm{SAB}(\mathrm{PMA}_k(\mathbf{Z})))\in\mathbb{R}^{k\times d},\end{equation}
\begin{equation}
	\text{where}\ \  \text{PMA}_k(\mathbf{Z}) := \text{MAB}(\mathbf{S}, \mathrm{rFF}(\mathbf{Z}))\in\mathbb{R}^{k\times d}.
\end{equation}

The SIHA-specific HSP-sa encoder--decoder mapping is given in Section~\ref{sec:hsp_sa}. HSP-sg retains this topology and applies the structural biases described in Section~\ref{sec:method} to the encoder self-attention, PMA, and decoder self-attention blocks.

\section{Training and Reproducibility Details}
\label{training_details}
Algorithm~\ref{alg:training} specifies the cache-based HSP-sg training procedure. This section reports the configurations used for (1) separate pretraining of HSP-sa and NRI, (2) iterative training of HSP-sg, and (3) deployment-time refinement.

\subsection{Pretraining of HSP-sa and NRI on Fully Observed Data}

HSP-sa and NRI are pretrained separately on complete trajectories. For HSP-sa, a subset of agents is masked from the input and its ground-truth trajectories are used as supervised reconstruction targets. The synthetic systems use Hungarian-aligned MSE, whereas the motion-capture experiments use direct joint-wise MSE in the predefined masked-joint order. NRI receives the complete trajectories and is optimized with the standard NRI objective, without ground-truth interaction labels.

For the synthetic systems, the HSP-sa module is implemented as a Set Transformer with a hidden dimension of \texttt{256} and head number of \texttt{4}. It is optimized using the Adam optimizer with a learning rate of \texttt{5e-4}, a weight decay of \texttt{1e-6} for Springs and \texttt{1e-5} for Charged Particles and Kuramoto, and batch size of \texttt{128}. Training is performed for \texttt{500} epochs.

For the synthetic systems, the NRI module adopts the standard encoder-decoder architecture with Gumbel-Softmax edge sampling. The number of edge types is fixed to \texttt{2}. It is optimized using the Adam optimizer with a learning rate of \texttt{5e-4}, and batch size of \texttt{128}. Training is performed for \texttt{200} epochs. The pretrained models are then used to generate initial structure estimates and hidden state predictions for the iterative training stage.

\subsection{Training of Structure-Guided Predictor (HSP-sg)}

The structure-guided hidden-state predictor (HSP-sg) is trained using structure estimates initialized by the pretrained HSP-sa and NRI modules. The predicted structure is stored in a per-sample cache and periodically recomputed by the pretrained NRI module. During this stage, HSP-sg is optimized with the dataset-appropriate hidden-state reconstruction loss, while the pretrained modules provide the initialization and structure updates. For the synthetic systems, this is Hungarian-aligned MSE; for motion capture, it is direct joint-wise MSE in the predefined masked-joint order. Algorithm~\ref{alg:training} gives the common cache-based training procedure.

\begin{algorithm}[h]

\caption{Iterative Training of HSP-sg}
\begin{algorithmic}[1]
\label{alg:training}
\STATE \textbf{Input:} Training dataset $\{x_\text{vis}, x_\text{hid}\}$, pre-trained $f_\text{pre}$ and $g_\text{NRI}$
\STATE \textbf{Output:} Trained $f_\text{sg}$ and updated structure cache
\STATE Initialize structure cache $\mathbf{A}_{\text{cache}}$ with $g_\text{NRI}([x_\text{vis},f_\text{pre}(x_\text{vis})])$
\FOR{epoch $=1$ to $E$}
    \FOR{each mini-batch in training data}
        \STATE Fetch cached structure $\mathbf{A}_{\text{cache}}$
        \STATE Use $f_\text{sg}$ with structure guidance $\mathbf{A}_{\text{cache}}$ to predict hidden states $\hat{x}_\text{hid}$
        \STATE Compute the dataset-appropriate hidden-state reconstruction loss
        \STATE Update parameters of $f_\text{sg}$ via backpropagation
    \ENDFOR
    \IF{epoch $> T_0$ \AND epoch $\bmod~M = 0$}
        \STATE Update $\mathbf{A}_{\text{cache}}$ using $g_\text{NRI}([x_\text{vis},f_\text{sg}(x_\text{vis},\mathbf{A}_{\text{cache}})])$
    \ENDIF
\ENDFOR
\end{algorithmic}
\end{algorithm}

During each synthetic training phase, HSP-sg receives the visible trajectories and current structure cache as input. The predicted hidden states are combined with the visible ones and passed to the NRI model to update the structure estimates, which are written back to the cache every $10$ epochs after warm-up cache freezing for the first \texttt{80} epochs. 

For the synthetic systems, we train HSP-sg using the Adam optimizer with a learning rate of \texttt{5e-4}, a weight decay of \texttt{1e-6} for Springs and \texttt{1e-5} for Charged Particles and Kuramoto, batch size of \texttt{128}. Each full training run consists of \texttt{500} epochs, and MSE is computed after aligning predicted hidden states to ground truth via Hungarian matching.

\subsection{Evaluation-time Refinement via Iterative Structure Cache}

At test time, the structure cache is initialized using the pretrained models and refined for \texttt{5} update rounds. In each round, the HSP-sg model is used to predict hidden states, followed by a forward pass through the NRI module to update structure predictions.

\subsection{System-specific Settings and Embedding Modules}

For the Kuramoto system, we use a three-dimensional state representation consisting of phase difference, amplitude, and intrinsic frequency. We also fix the null interaction type as always inactive to reflect the system's continuous coupling nature.

All systems use Set Transformer-based architectures for the HSP modules. For trajectory embedding, Springs uses an MLP-based embedding, whereas Charged Particles and Kuramoto use one-dimensional convolutional embeddings. These system-specific choices follow the corresponding NRI configurations to maintain architectural consistency with the NRI baselines.

\subsection{Motion-Capture Training Configuration}

For both motion-capture subjects, HSP-sa and HSP-sg use a hidden dimension of \texttt{256}, \texttt{4} attention heads, dropout of \texttt{0}, a history length of \texttt{49}, and an input dimension of \texttt{6} corresponding to 3D position and 3D velocity. With random seed \texttt{1}, the HSP modules are trained for \texttt{500} epochs using a batch size of \texttt{8}, a learning rate of \texttt{1e-4}, and weight decay of \texttt{1e-6}. Their outputs follow the predefined masked-joint order, and training uses direct joint-wise MSE. For HSP-sg, the structure cache has a warm-up of \texttt{40} epochs and is updated every \texttt{20} epochs; interaction guidance uses edge type index \texttt{1}.

The motion-capture backbone is a static-graph NRI model with \texttt{2} edge types, an encoder hidden dimension of \texttt{256}, an encoder MLP hidden dimension of \texttt{256} with \texttt{3} layers, and a decoder hidden dimension of \texttt{256}. It uses \texttt{skip\_first=true}, a Gumbel-Softmax temperature of \texttt{0.5}, and \texttt{10} teacher-forcing steps. Each subject-specific NRI model is trained for \texttt{500} epochs using a batch size of \texttt{8} and a learning rate of \texttt{5e-4}.

\subsection{Dataset Generation and Scale}

We use the Springs, Charged Particles, and Kuramoto systems as synthetic benchmark environments, following the trajectory-based evaluation setting used in NRI~\cite{kipf2018neural}. Each trajectory contains $T=50$ time steps for training and validation and $T=100$ time steps for testing. Springs and Charged Particles use four-dimensional position--velocity states, whereas Kuramoto uses the three-dimensional representation consisting of phase difference, amplitude, and intrinsic frequency.

We use 200,000 samples for training, 50,000 for validation, and 50,000 for testing. These dataset sizes exceed those used in the original NRI experiments and are reported here as part of our experimental configuration.

\subsection{Compute Resources}

All experiments were conducted on a single NVIDIA RTX 4090 GPU with 24 GB memory using PyTorch 2.6.0 and CUDA 12.6. A complete training run, including pretraining and iterative refinement, takes approximately 5--15 hours for systems with 5--10 agents. Training scripts, data-generation tools, and configuration files will be released publicly.

\section{Complete Quantitative Results}
\label{app:quantitative_results}

\subsection{Variants without Hidden-State Supervision}
\label{d3}
We further study whether the hidden-state predictor can be trained without direct supervision on hidden states.
We consider three variants:
(1) jointly training NRI and HSP-sa from scratch, with both modules randomly initialized;
(2) jointly training NRI and HSP-sa while initializing NRI from a model pretrained on fully observed data; and
(3) training only HSP-sa while keeping an NRI model pretrained on fully observed data frozen.
In all three cases, supervision is provided only through the future states of the visible agents.
That is, HSP-sa is not directly supervised by ground-truth hidden trajectories, but remains in the end-to-end backpropagation chain through the visible future prediction loss.
Variant (1) uses neither fully observed NRI pretraining nor direct hidden-state supervision. For reference, we also report the main supervised setting, in which ground-truth trajectories of the masked agents supervise HSP-sa reconstruction during training.

Table~\ref{tab:no_hidden_supervision} summarizes the results on the Springs task with $N_{\text{vis}}=5$ and $N_{\text{hid}} \in \{1,2,3,4,5\}$.
All three variants without hidden-state supervision have substantially higher hidden-state and visible-future errors than the supervised HSP-sa setting, and their visible-to-visible structural accuracy also deteriorates as $N_{\text{hid}}$ increases. These results characterize the behavior of the three tested visible-future-only training objectives; they do not constitute an impossibility result for other objectives or models. The main experiments therefore retain direct hidden-state supervision during training, as specified in Section~\ref{sec:problem_formulation}.
\begin{table*}[h]
\centering
\caption{Results of variants without hidden-state supervision on the Springs task with $N_{\text{vis}}=5$ and $N_{\text{hid}}\in\{1,2,3,4,5\}$. Results are in the format of $\mathrm{MSE}_{\mathrm{HSP}}$ / $\mathrm{MSE}_{\mathrm{FSP,Vis.}}$ / $\mathrm{ACC}_{\mathrm{V\mbox{-}V}}$. Lower MSE and higher ACC are better.}
\label{tab:no_hidden_supervision}
\resizebox{\textwidth}{!}{
\begin{tabular}{l|ccccc}
\toprule
Method 
& $N_{\text{hid}}=1$ 
& $N_{\text{hid}}=2$ 
& $N_{\text{hid}}=3$ 
& $N_{\text{hid}}=4$ 
& $N_{\text{hid}}=5$ \\
\midrule
Joint train NRI + HSP-sa (random init) 
& 4.3e-1 / 2.0e-4 / 98.2 
& 3.6e-1 / 1.8e-4 / 72.1 
& 2.8e-1 / 2.3e-4 / 65.7 
& 3.3e-1 / 2.3e-4 / 50.0 
& 4.0e-1 / 2.8e-4 / 50.0 \\

Joint train NRI + HSP-sa (pretrained NRI) 
& 4.1e-1 / 2.2e-4 / 98.4 
& 3.3e-1 / 2.3e-4 / 73.5 
& 2.6e-1 / 2.4e-4 / 66.2 
& 3.0e-1 / 2.6e-4 / 50.0 
& 3.7e-1 / 2.9e-4 / 50.0 \\

Train HSP-sa only + freeze pretrained NRI 
& 3.8e-1 / 2.2e-4 / 98.2 
& 3.1e-1 / 2.4e-4 / 73.8 
& 2.5e-1 / 2.4e-4 / 66.1 
& 2.7e-1 / 2.8e-4 / 50.0 
& 3.5e-1 / 3.1e-4 / 50.0 \\

Supervised HSP-sa on fully observed data 
& 2.5e-3 / 1.5e-5 / 99.7 
& 5.6e-3 / 9.3e-6 / 98.7 
& 7.1e-3 / 6.4e-6 / 97.0 
& 8.1e-3 / 5.7e-6 / 95.6 
& 9.1e-3 / 6.0e-6 / 93.0 \\
\bottomrule
\end{tabular}
}
\end{table*}

\subsection{Ablation on Encoder Modules for HSP-sa}
\label{app:encoder_ablation}

We perform an ablation study to evaluate the effect of different encoder architectures used in the structure-agnostic hidden state predictor (HSP-sa). Specifically, we replace the Set Transformer encoder with alternative modules including a standard MLP, vanilla Transformer, GNN, and Slot Attention~\cite{locatello2020object} encoder. All variants are trained on the Springs dataset with $N_{\text{vis}} = 5$ visible agents and $N_{\text{hid}} = 2$ hidden agents, under the same training configuration and data size.

Table~\ref{tab:encoder-ablation} reports the hidden-state prediction MSE for each encoder choice. The Set Transformer has the lowest reported MSE among these variants; this table does not by itself isolate which architectural property accounts for the difference.

\begin{table}[h]
\centering
\caption{Hidden state prediction MSE for HSP-sa with different encoder modules on the Springs dataset ($N_{\text{vis}}=5$, $N_{\text{hid}}=2$).}
\label{tab:encoder-ablation}
\vspace{0.5em}
\begin{tabular}{lc}
\toprule
\textbf{Encoder Module} & \textbf{MSE$_{\text{HSP}}$} \\
\midrule
MLP &  $1.2 \times 10^{-2}$ \\
Transformer &  $1.1 \times 10^{-2}$ \\
GNN &  $1.6 \times 10^{-2}$ \\
Slot Attention &  $1.6 \times 10^{-2}$ \\
\textbf{Set Transformer} &  $\mathbf{5.6 \times 10^{-3}}$ \\
\bottomrule
\end{tabular}
\end{table}

\subsection{Repeated-run statistics on representative settings}
\label{app:seed_stats}

To assess statistical stability, we report repeated-run results for three representative settings: Springs, Charged Particles, and Kuramoto with $N_{\rm vis}=5$ and $N_{\rm hid}=3$. All models, including SIHA (HSP-sg) and the pretrained NRI and HSP-sa modules, are trained independently over five random seeds.

\begin{table}[h]
\centering
\caption{Five-run statistics on representative settings. We report mean $\pm$ standard deviation over 5 independent runs with different random seeds.}
\label{tab:seed_stats_main}
\resizebox{\textwidth}{!}{
\begin{tabular}{llcccccc}
\toprule
Dataset & Method 
& $\mathrm{MSE}_{\mathrm{HSP}} \downarrow$
& $\mathrm{MSE}_{\mathrm{FSP,Vis.}} \downarrow$
& $\mathrm{MSE}_{\mathrm{FSP,Hid.}} \downarrow$
& $\mathrm{ACC}_{\mathrm{V\mbox{-}V}} \uparrow$
& $\mathrm{ACC}_{\mathrm{V\mbox{-}H}} \uparrow$
& $\mathrm{ACC}_{\mathrm{H\mbox{-}H}} \uparrow$ \\
\midrule

\multirow{3}{*}{Springs}
& NRI
& --
& $3.7\mathrm{e}{-5} \pm 2.3\mathrm{e}{-6}$
& --
& $72.9 \pm 0.2$
& --
& -- \\
& HSP-sa
& $7.1\mathrm{e}{-3} \pm 4.0\mathrm{e}{-5}$
& $6.3\mathrm{e}{-6} \pm 3.4\mathrm{e}{-7}$
& $1.3\mathrm{e}{-2} \pm 1.7\mathrm{e}{-3}$
& $97.0 \pm 0.4$
& $83.4 \pm 0.8$
& $61.3 \pm 1.1$ \\
& SIHA (HSP-sg)
& $\mathbf{4.9\mathrm{e}{-3} \pm 6.7\mathrm{e}{-5}}$
& $\mathbf{5.0\mathrm{e}{-6} \pm 3.1\mathrm{e}{-7}}$
& $\mathbf{1.0\mathrm{e}{-2} \pm 1.6\mathrm{e}{-3}}$
& $\mathbf{97.2 \pm 0.4}$
& $\mathbf{88.7 \pm 0.6}$
& $\mathbf{67.6 \pm 1.0}$ \\
\midrule

\multirow{3}{*}{Charged}
& NRI
& --
& $\mathbf{2.5\mathrm{e}{-3} \pm 1.7\mathrm{e}{-4}}$
& --
& $62.1 \pm 0.4$
& --
& -- \\
& HSP-sa
& $4.3\mathrm{e}{-2} \pm 2.0\mathrm{e}{-4}$
& $2.6\mathrm{e}{-3} \pm 2.2\mathrm{e}{-4}$
& $1.3\mathrm{e}{-1} \pm 1.8\mathrm{e}{-2}$
& $71.3 \pm 0.7$
& $57.5 \pm 2.2$
& $\mathbf{52.1 \pm 2.4}$ \\
& SIHA (HSP-sg)
& $\mathbf{4.1\mathrm{e}{-2} \pm 3.8\mathrm{e}{-4}}$
& $2.6\mathrm{e}{-3} \pm 2.3\mathrm{e}{-4}$
& $\mathbf{1.2\mathrm{e}{-1} \pm 1.7\mathrm{e}{-2}}$
& $\mathbf{72.4 \pm 0.8}$
& $\mathbf{58.9 \pm 2.5}$
& $\mathbf{52.1 \pm 2.7}$ \\
\midrule

\multirow{3}{*}{Kuramoto}
& NRI
& --
& $4.2\mathrm{e}{-2} \pm 1.9\mathrm{e}{-3}$
& --
& $65.2 \pm 0.3$
& --
& -- \\
& HSP-sa
& $1.1\mathrm{e}{-1} \pm 6.0\mathrm{e}{-4}$
& $1.6\mathrm{e}{-2} \pm 1.2\mathrm{e}{-3}$
& $1.3\mathrm{e}{-1} \pm 1.8\mathrm{e}{-2}$
& $83.6 \pm 0.5$
& $68.6 \pm 0.7$
& $53.3 \pm 1.8$ \\
& SIHA (HSP-sg)
& $\mathbf{1.0\mathrm{e}{-1} \pm 6.3\mathrm{e}{-4}}$
& $\mathbf{1.5\mathrm{e}{-2} \pm 1.3\mathrm{e}{-3}}$
& $\mathbf{1.2\mathrm{e}{-1} \pm 1.6\mathrm{e}{-2}}$
& $\mathbf{84.0 \pm 0.7}$
& $\mathbf{70.2 \pm 0.9}$
& $\mathbf{54.8 \pm 2.0}$ \\
\bottomrule
\end{tabular}
}
\end{table}

\section{Additional Analyses and Motion-Capture Details}
\label{app:additional_analyses}

\subsection{Motion-Capture Focus-Graph Construction}
\label{app:mocap_details}

The focus plots in Figure~\ref{fig:mocap_six_panel} are constructed around the left or right hand. For a focus joint $i$, the score of joint $j$ is $s_j=(A_{ij}+A_{ji})/2$, with $s_i=0$. The visualization retains the top-$k$ edges with score greater than $0.75$: panels (a) and (b) use $k=8$, whereas panels (c)--(f) use $k=14$. The full-observation NRI structure is shown only as a reference estimate and is not treated as a ground-truth graph.

\subsection{Unknown and Variable Numbers of Hidden Agents}
\label{d1}

We evaluate an existing model-selection procedure when the number of hidden agents is unknown and takes a value in $N_{\text{hid}}\in\{0,\dots,5\}$. HSP-sa and NRI models configured for different $N_{\rm hid}$ values are applied separately, and the selected count is the one yielding the lowest visible-agent future-prediction MSE. Table~\ref{tab:nhid_acc} reports the resulting count-prediction accuracy.

The accuracy remains above $83\%$ for all reported Springs settings, but decreases from $92.3$ to $45.5$ on Charged and from $87.5$ to $24.7$ on Kuramoto as the true count changes from $0$ to $5$. The result is therefore treated as a boundary analysis rather than evidence that the current fixed-count formulation resolves unknown cardinality.

\begin{table}[h!]
\centering
\caption{Accuracy (\%) of hidden-agent count prediction across different numbers of hidden agents $N_{\text{hid}}$.}
\small
\begin{tabular}{lcccccc}
\toprule
& \multicolumn{6}{c}{$N_\text{hid}$} \\
\cmidrule(lr){2-7}
Dataset & 0 & 1 & 2 & 3 & 4 & 5 \\
\midrule
Springs & 99.9 & 99.1 & 90.5 & 88.8 & 87.2 & 83.7 \\
Charged & 92.3 & 88.2 & 76.5 & 69.1 & 56.2 & 45.5 \\
Kuramoto & 87.5 & 67.0 & 49.5 & 36.7 & 28.4 & 24.7 \\
\bottomrule
\end{tabular}
\label{tab:nhid_acc}
\end{table}

\subsection{Additional Unmodeled Hidden Agents}
\label{d2}

We additionally evaluate a Springs setting with $N_{\rm hid}=2$ and $N_{\rm vis}=5$, in which $N_{\rm add}\in\{1,2,3\}$ additional agents are hidden from the model while it is trained to reconstruct only the first two hidden trajectories from the five visible trajectories.

Table~\ref{tab:robustness} shows progressive degradation in the three displayed metrics as $N_{\text{add}}$ increases. This experiment documents sensitivity to additional unmodeled agents; it is not used to claim general robustness to arbitrary hidden-agent configurations.

\begin{table}[h]
\centering
\caption{SIHA (HSP-sg) results on Springs when the data contain additional unmodeled hidden agents. For comparison, the NRI reference has an $\text{MSE}_{\text{FSP, Vis.}}$ of 3.0e-5 and an $\text{ACC}_{\text{V-V}}$ of 76.1.}
\begin{tabular}{lcccc}
\toprule
& \multicolumn{4}{c}{$N_\text{add}$} \\
\cmidrule(lr){2-5}
& 0 & 1 & 2 & 3 \\
\midrule
$\text{MSE}_{\text{HSP}}$     & 4.9e-3 & 1.2e-2 & 1.3e-2 & 1.5e-2 \\
$\text{MSE}_{\text{FSP, Vis.}}$   & 5.0e-6 & 7.1e-6 & 7.9e-6 & 9.0e-6 \\
$\text{ACC}_{\text{V-V}}$    & 97.2 & 92.3 & 85.8 & 77.1 \\
\bottomrule
\end{tabular}
\label{tab:robustness}
\end{table}

\end{document}